\documentclass[9pt,twocolumn,twoside]{opticajnl}
\journal{arxiv}
\setboolean{shortarticle}{false}

\usepackage{hyphenat}
\usepackage{graphicx} 
\usepackage{bm}
\usepackage{stfloats}
\usepackage{xcolor}
\usepackage{pifont}

\definecolor{capgreen}{RGB}{0,140,60}
\definecolor{capamber}{RGB}{215,145,0}
\definecolor{capred}{RGB}{200,30,30}

\newcommand{\cyes}{\textcolor{capgreen}{\ding{51}}}
\newcommand{\cpart}{\textcolor{capamber}{$\sim$}}
\newcommand{\cno}{\textcolor{capred}{\ding{55}}}

\title{Crazyflow: An Accurate, GPU-Accelerated, Differentiable Drone Simulator in JAX}

\author[$\ast$,1, $\dagger$]{Martin~Schuck}
\author[$\ast$, 1]{Marcel~P.~Rath}
\author[1]{Yufei~Hua}
\author[2]{Abhishek~Goudar}
\author[3]{SiQi~Zhou}
\author[1]{Angela~P.~Schoellig}

\affil[$\ast$]{Equal contribution.}
\affil[1]{Technical University of Munich}
\affil[2]{University of Toronto}
\affil[3]{Simon Fraser University}
\affil[$\dagger$]{Corresponding author. Email: martin.schuck[at]tum.de}

\date{May 2026}

\setboolean{displaycopyright}{false} 

\begin{abstract}
High-quality, large-scale synthetic data from simulations is becoming a cornerstone for pushing robots' capabilities. While aerial robotics simulators have evolved to support specialized needs such as fidelity, differentiability, and swarms independently, a unified platform that synthesizes data across all these domains is missing. In this work, we propose \textsc{Crazyflow}, a simulator designed to push the frontier of aerial robotics algorithm development, supporting model-based and data-driven methods, gradient-based and sampling-based approaches, single- and multi-agent systems, and state- and vision-based policy learning. \textsc{Crazyflow} combines high-fidelity dynamics with JAX for speed and differentiability, scaling to millions of environments or thousands of swarms with thousands of drones on single CPUs and GPUs, and beyond on clusters. Real-world experiments show \textsc{Crazyflow} supports both analytical-gradient-based policy learning, achieving high-accuracy tracking without domain randomization, and sampling-based obstacle avoidance at speeds exceeding half a billion steps per second. Breaking the traditional train-then-deploy paradigm, \textsc{Crazyflow}'s performance enables in-flight reinforcement learning; we demonstrate this by throwing a physical drone into the air and training a recovery policy from scratch in 0.37 seconds, catching it before it hits the ground. \textsc{Crazyflow} supports multiple levels of simulation abstraction, provides high-performance rendering for vision-based training, is compatible with all open-source Crazyflie models, and enables rapid reconfiguration across custom drone platforms by providing a lightweight system identification pipeline. By pushing accuracy, speed, and differentiability simultaneously, \textsc{Crazyflow} serves as an open-source resource for synthetic data generation, with emerging capabilities for large-scale parallelization for online, in-execution learning and optimization, opening the door to novel algorithm development.
\noindent
\url{https://learnsyslab.github.io/crazyflow}
\end{abstract}

\begin{document}

\maketitle
\sloppy

\section*{Introduction}
\noindent
Owing to their high agility, mechanical simplicity, and cost-effective design, drones have emerged as a canonical platform across a broad spectrum of robotics applications, ranging from safety-critical tasks such as search and rescue, infrastructure inspection, and construction monitoring~\cite{lyu2023unmanned,guan2022review} to recreational and creative domains such as autonomous racing, aerial cinematography, and large-scale drone performances~\cite{hanover2024autonomous,mademlis2019highlevel,schoellig2014so}. This broad adoption has, in turn, catalyzed research efforts toward two frontiers. One focuses on reliably pushing individual drones toward their physical limits, while the other focuses on scaling up multi-agent systems for complex, collective tasks. Advancing capabilities along either frontier necessitates high-performance, computationally efficient tools that enable rapid algorithmic iteration to fully exploit the drones' unique maneuverability within three-dimensional, and possibly unstructured, environments.

\begin{figure*}[t]
    \centering
    \includegraphics[width=2\columnwidth]{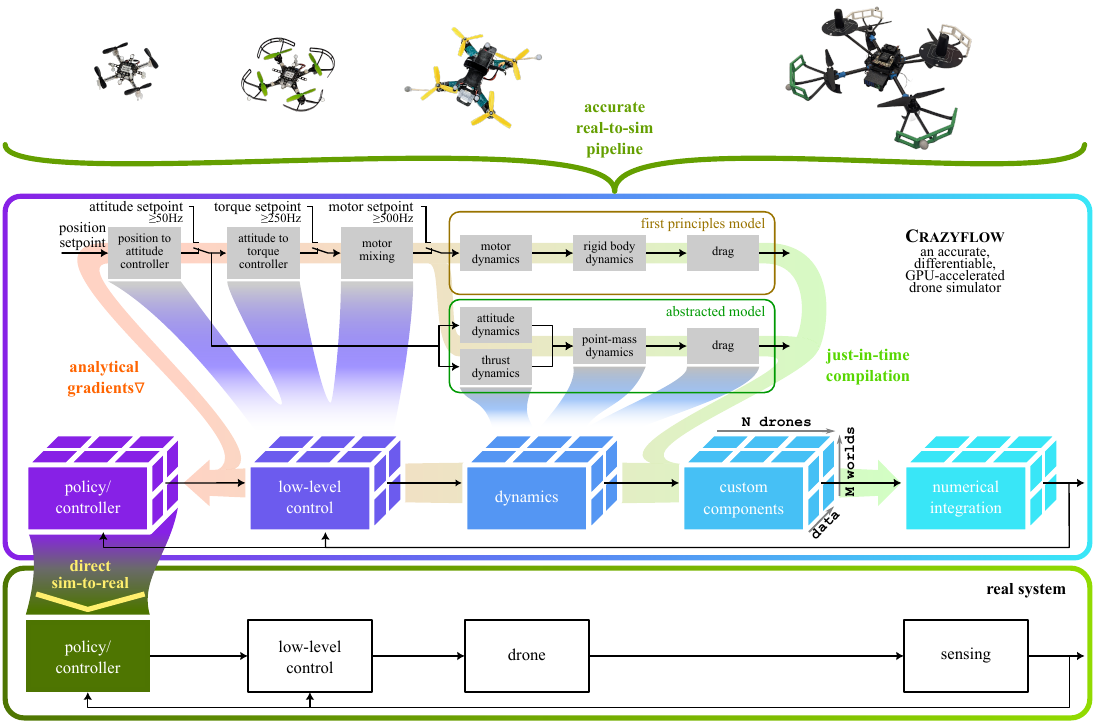}
    \caption{\textbf{An overview of the \textsc{Crazyflow} simulator.} Built on \textsc{JAX}, \textsc{Crazyflow} leverages just-in-time compilation via Accelerated Linear Algebra (XLA) to unify physics and controllers into a single differentiable computation graph. The simulation has been built from the ground up for parallelization and swarms: data passed between steps is represented as $M_{worlds} \times N_{drones}$- sized tensors, visualized here as 3D cubes. Our modular pipeline supports two types of models, one from first principles and an abstracted model identified from flight data, with reliable sim-to-real transfer for the Crazyflie family of drones and customized platforms. Website and videos: \protect\url{https://learnsyslab.github.io/crazyflow}}
    \label{fig:overview}
\end{figure*}

Among the software tools required to support these advances, simulators have become an indispensable cornerstone in the development of aerial robotics autonomy stacks. High-performance simulators enable rapid prototyping of theoretical frameworks, systematic tuning of algorithmic parameters, and stress-testing of methods against potential failure modes in a risk-free, low-cost environment. Recent breakthroughs in deep reinforcement learning (RL) have further underscored the importance of simulation, demonstrating that massive parallelization is essential for solving non-trivial control problems that would be impractical to explore directly on hardware~\cite{rudin2022learning, kaufmann2023champion}. Crucially, however, this computational throughput must be grounded in physical accuracy to ensure that behavior learned or optimized in simulation can transfer reliably to real-world platforms~\cite{hwangbo2019learning, kaufmann2023champion, sorocky2020experience}. Complementing this trend is the emergence of differentiable simulation, which enables algorithms to exploit exact analytic gradients for efficient, gradient-based policy optimization in both data-driven and model-based settings~\cite{newbury2024review}. Ultimately, the utility of a modern simulator is increasingly dependent on three properties: execution speed, physical accuracy, and differentiability. These factors are critical for accelerating algorithm design cycles and generating inexpensive, high-quality synthetic data to build confidence before large-scale real-world deployment.

Over the past decade, a diverse set of simulation tools has been developed for drone research, prioritizing high-fidelity modeling and validation of autonomy pipelines. Commonly used simulators such as  RotorS~\cite{furrer2016rotors}, CrazyS~\cite{silano2018crazys}, and CrazySim~\cite{llanes2024crazysim}, along with developments such as sim\_cf~\cite{sim_cf2018}, are built on modular frameworks and provide high-fidelity models suitable for controller development and system-level testing, with the possibility of software-in-the-loop or even hardware-in-the-loop evaluations. While these are valuable resources for validating controllers prior to deployment, their CPU-based architectures are not designed for high-throughput parallel simulation, making them less suitable for large-scale learning or extensive algorithmic parameter optimization. Additional frameworks such as Aerostack2~\cite{fernandez2023aerostack2}, CrazySwarm~\cite{preiss2017crazyswarm}, and CrazyChoir~\cite{pichierri2023crazychoir} provide excellent modular features tailored for multi-agent deployment; however, they similarly lack native support for high-throughput batched simulation typically required by large-scale learning-oriented methods.

To address the computational demands of large-scale learning, recent work has pivoted toward massively parallel simulation, leveraging different acceleration strategies. For instance, Aerial Gym~\cite{kulkarni2025aerialgym} builds on the GPU-native NVIDIA Isaac Gym engine to achieve high simulation throughput via tensorized physics~\cite{makoviychuk2021isaacgym}, PRL4AirSim~\cite{saunders2023parallel} leverages a distributed computing architecture to scale AirSim~\cite{shah2017airsim} instances across networked clusters, and gym\_pybullet\_drones~\cite{panerati2021learning} employs process-level parallelism to run multiple PyBullet physics engines~\cite{coumans2016pybullet} concurrently on a single machine. Moreover, frameworks such as QuadSwarm~\cite{huang2023quadswarm} focus on scaling up multi-agent RL by leveraging optimized, ad-hoc dynamics modules, while Flightmare~\cite{song2021flightmare} and Agilicious~\cite{foehn2022agilicious} achieve high simulation throughput by further decoupling photorealistic rendering and sensor simulation from physics, combined with lightweight C++ dynamics implementations. In a similar spirit, Learning to Fly (L2F)~\cite{eschmann2024learning_to_fly_in_seconds} focuses exclusively on scalable drone dynamics and implements parallelized environments in C++ and CUDA. While these approaches offer substantial speedups and improved scalability, they typically preclude the use of analytical gradients required for advanced learning methods, e.g.,~\cite{xu2022accelerated,sun2025flowmatching}.

Emerging frameworks attempt to bridge this gap by leveraging automatic differentiation. DiffPhysDrone~\cite{zhang2025visionbased} implements dynamics using \textsc{PyTorch}~\cite{paszke2019pytorch} to enable end-to-end learning. However, its backend simulation relies primarily on simplified point-mass models (i.e., double integrators) with customized acceleration-level control interfaces. This design, tailored to specific platforms, implicitly assumes that the downstream low-level controller can reliably track desired accelerations, thereby offloading the complexity of sim-to-real transfer rather than capturing it within the physics loop. DiffAero~\cite{zhang2025diffaero} further incorporates multiple levels of model fidelity, ranging from point-mass dynamics to idealized rigid-body drone models. Aside from built-in support for automatic differentiation, \textsc{PyTorch} is more flexible compared to specialized C++ implementations. This flexibility is one of the fundamental drivers behind the explosion in deep learning research; however, in high-performance tasks such as robotics simulations, the inefficiencies of such frameworks also quickly become evident.

While frameworks like \textsc{PyTorch} delegate computations to optimized backend kernels, they fundamentally operate on an eager execution model~\cite{paszke2019pytorch}. This means each operation is executed line by line, necessitating frequent synchronization between the hardware accelerators and the Python interpreter. To overcome these limitations, emerging frameworks such as \textsc{JAX}~\cite{jax2018github} have introduced a paradigm shift toward lazy execution with just-in-time (JIT) compilation. Here, computations are first traced without execution and then fused into a single, fully-optimized function using the XLA compiler~\cite{sabne2020xla}. This architecture eliminates unnecessary synchronization steps from the resulting computation graph and optimizes gradient computations, yielding significant speedups compared to eager frameworks. The transition towards lazy, compiled computation graphs represents a significant trend in deep learning and optimization research~\cite{rutherford2024jaxmarl, chalumeau2024qdax, kwon2026volumetric, kidger2021equinox, suarez2025pufferlib}. In robotic simulations, new engines and frameworks such as Brax~\cite{freeman2021brax}, MuJoCo XLA (MJX)~\cite{deepmind2026mjx}, NVIDIA Warp~\cite{macklin2022warp}, and Newton~\cite{contrib2025newton} are adopting this paradigm, bridging the long-standing gap between the modularity of high-level languages and the raw performance of native C++ implementations.

In this work, we introduce \textsc{Crazyflow}, a high-fidelity, differentiable simulator built on \textsc{JAX}. To enable accurate physics simulation down to the actuator level while supporting different control interfaces, we identify necessary effects to reliably model the drones' behavior and emulate a standard onboard controller stack for JAX's compilation and differentiation. This stack is directly compatible with Crazyflies, a family of widely used open-source platforms for drone research. To support other platforms, we offer a streamlined system identification pipeline that adapts models to new drones with only minutes of flight data. To support model-based control applications, all physics models are also available in symbolic form via CasADi~\cite{andersson2019casadi}.

\textsc{Crazyflow}'s architecture (see \autoref{fig:overview} and Materials and Methods, ``Simulator Design'') is built to support massive parallelization for single drones and large-scale swarms at full fidelity. Aligning with state-of-the-art general-purpose engines such as MJX, our framework leverages XLA to unify physics and controllers into a single differentiable computation graph. \textsc{JAX}'s tracing makes our simulation pipeline highly flexible, allowing researchers to compile custom components, disturbances, and additional aerodynamic effects directly into the computation graph without adding any execution overhead (Materials and Methods, ``Customizing the Simulation''). This flexibility also extends to vision-based pipelines through JAX-compatible photorealistic rendering based on Gaussian splat representations. Specializing in drone flight allows us to avoid the costly solve step of general-purpose simulators, yielding substantial additional speedups while also offering high-fidelity drone physics that enable direct sim-to-real transfer even in the absence of domain randomization. A detailed comparison of \textsc{Crazyflow}'s control stack, modeling, computation, differentiability, swarm capabilities, and sensing  against existing simulators is provided in \autoref{tab:sup:simulator-capabilities} of the Supplementary Text.

As we highlight in our results, by bridging accuracy, speed, and differentiability in a unified framework, \textsc{Crazyflow} provides a foundation for moving robot learning beyond the traditional train-then-deploy paradigm towards real-time, in-flight learning and optimization at scale. We show that combining a high-fidelity dynamics model with efficient computation enables the training of high-performance RL agents in seconds or even sub-seconds. In particular, with \textsc{Crazyflow}, we are able to train policies achieving up to sub-centimeter tracking accuracy in 1.56 seconds without any domain randomization, and solve throwing stabilization tasks in 0.37 seconds, allowing a policy to be trained in mid-air and deployed in time to catch the drone before it falls to the ground. On vision-based tasks, \textsc{Crazyflow}'s built-in Gaussian splat rendering features allow training end-to-end navigation policies that zero-shot transfer to the real world. \textsc{Crazyflow} scales to up to 4.2 million drones on a consumer-level GPU and over 900 million steps per second. Beyond deep RL, our simulator has been validated for various control- and learning-oriented applications, ranging from state-of-the-art optimal control and sampling-based control methods to swarm coordination frameworks. These results highlight \textsc{Crazyflow}'s support for multiple physics backends, enabling data-driven models to be leveraged across diverse application domains and algorithmic dimensions.

Our \textsc{Crazyflow} simulator is fully open-source, with the main package accessible at \url{https://github.com/learnsyslab/crazyflow} and installable via \texttt{pip install crazyflow}; detailed documentation is available at \url{https://learnsyslab.github.io/crazyflow}. To support the development of \textsc{Crazyflow} and the broader open science ecosystem, we have also contributed to foundational tools for differentiable 3D spatial computations as part of \textsc{SciPy} 1.17.0~\cite{2020SciPy-NMeth,schuck2025scipy} and integrated accurate dynamics models for the family of Crazyflie drones into the official firmware~\cite{crazyflie-firmware}. With the fully open-source, modular framework and thorough sim-to-real validation, we believe \textsc{Crazyflow} will serve as a valuable resource for accelerating research across multiple communities, including deep and multi-agent RL, model-based planning and control, differentiable programming, and beyond.

\section*{Results}
\begin{figure*}
    \centering
    \includegraphics[width=1.0\linewidth]{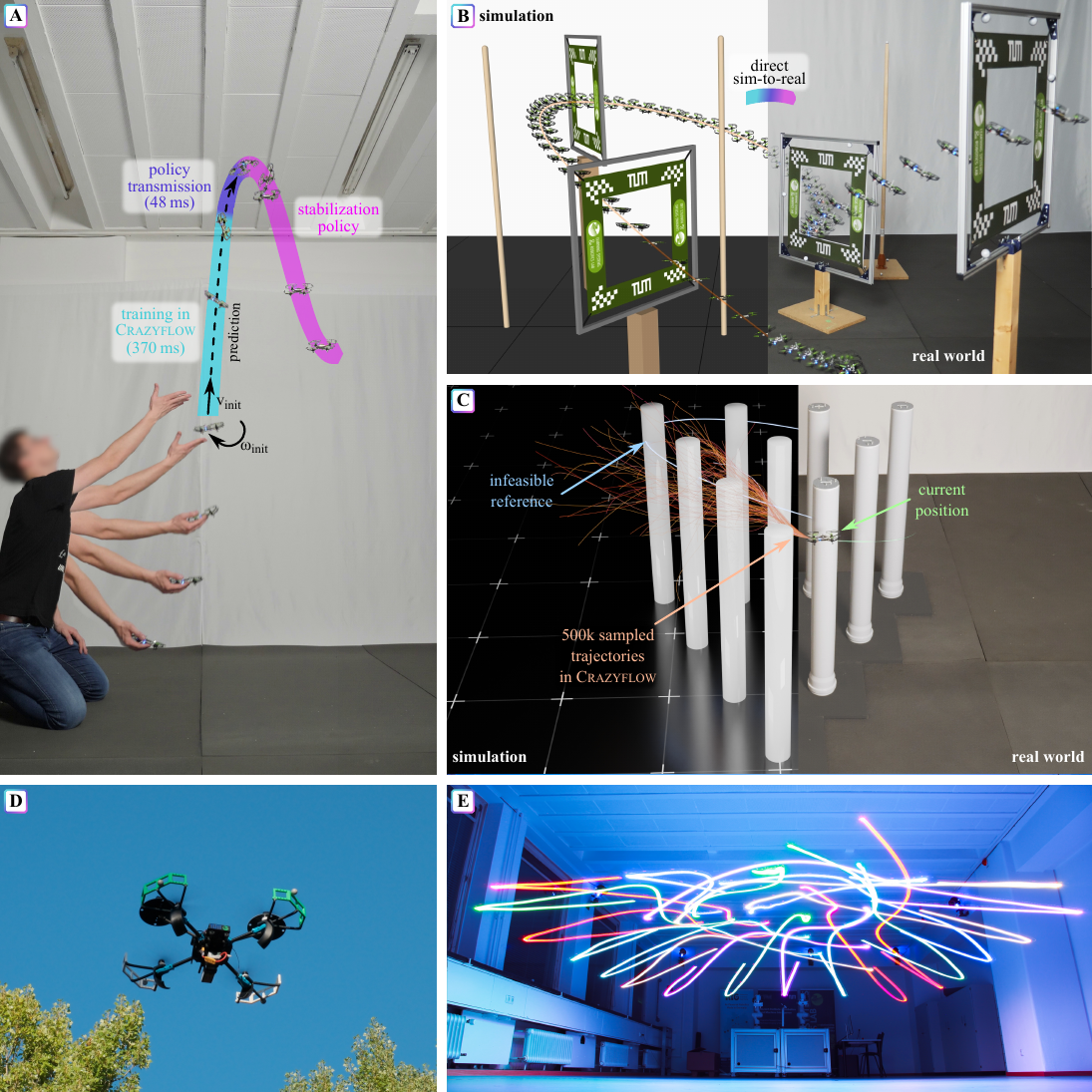}
    \caption{\textbf{Unifying diverse paradigms with \textsc{Crazyflow}.} Across a wide range of learning- and control-based approaches, \textsc{Crazyflow} allows for the fast development and easy deployment of new methods. We show this by training a stabilization policy based on the initial conditions in-flight (\textbf{A}), using it for academic courses on the graduate level (\textbf{B}), running a Model Predictive Path Integral (MPPI) controller natively (\textbf{C}), simulating drones for outdoor applications (\textbf{D}), and developing large-scale drone swarm applications (\textbf{E}) within our own group.}
    \label{fig:application}
\end{figure*}

\textsc{Crazyflow} is a versatile tool for aerial research that can be applied to a range of applications, from learning-oriented to control-oriented methods (\autoref{fig:application}\textbf{A}-\textbf{B}), from gradient-based to sampling-based algorithms (\autoref{fig:application}\textbf{C}), from standard open-source to customized platforms (\autoref{fig:application}\textbf{D}), and from single drones to swarms (\autoref{fig:application}\textbf{E}). In our experiments, we show that it seamlessly scales to millions of parallel environments and thousands of drones in a swarm, while maintaining a low sim-to-real gap across different platforms. \textsc{Crazyflow}'s speed, accuracy, and differentiability allow us to push algorithm limits, training high-performance, deployable policies in less than one second without domain randomization. The results presented in this section, including both simulation and real-time experiments, are summarized in our accompanying video (movies S1 -- S6). Details on the algorithms and parameters can be found in the Supplementary Text, and our experiments code.

\subsection*{Computation Performance} \label{sec:results:performance}

\begin{figure*}[t]
    \centering
    \includegraphics[width=0.99\textwidth]{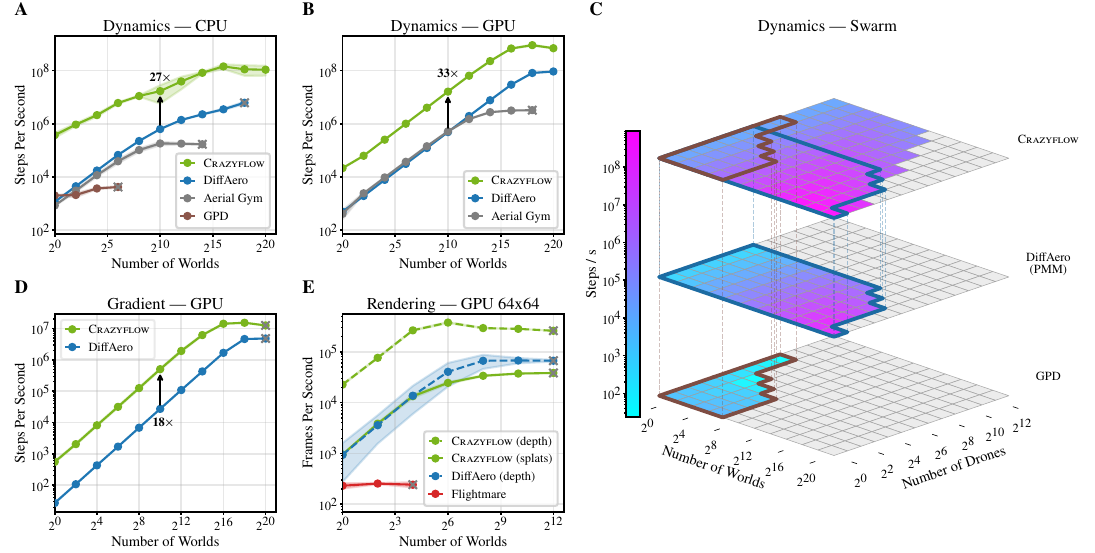}
    \vspace{-1em}
    \caption{\textbf{Performance and scaling evaluation of \textsc{Crazyflow}.} We show physics steps throughput per second for a single drone on the CPU (\textbf{A}) and the GPU (\textbf{B}) compared to other state-of-the-art simulators. We also compare the reachable configurations and throughput for combinations of parallel worlds and swarm sizes on the GPU against simulations with swarm support (\textbf{C}). Capability limits of prior works are projected onto \textsc{Crazyflow}'s heatmap. To measure throughput on gradient computations, we differentiate a goal distance with respect to the control inputs over ten simulation steps (\textbf{D}). Finally, for vision-based sensors, we compare both depth rendering with MJX against DiffAero (dashed lines) and full RGB-D rendering with Gaussian splatting against Flightmare (solid lines) (\textbf{E}). All data points are averaged over $50$ executions of $50$ steps including attitude controllers on the same machine. Missing data points exceed resource limits or cause crashes. Shaded intervals denote the confidence interval over 3 standard deviations. The full benchmarking protocol is provided in the Supplementary Text.}
    \label{fig:performance}
\end{figure*}

The combination of JIT compilation and vectorization allows \textsc{Crazyflow} to scale efficiently along two axes concurrently: we can simulate millions of drones across parallel worlds, swarms of thousands of drones, and even thousands of parallel worlds with thousands of drones, all using high-fidelity physics. On a consumer GPU, the simulator achieves a throughput of approximately 900 million steps per second at its peak. Compared to its predecessor, gym\_pybullet\_drones (GPD), \textsc{Crazyflow} improves throughput by two orders of magnitude at the same parallelization levels, and supports five orders of magnitude more environments while simulating at higher fidelity (see \autoref{fig:performance}\textbf{A}). As expected, CPU performance surpasses the GPU on a small number of parallel simulations, while GPU performance dominates after $2^{10}$ worlds

Because \textsc{JAX} statically traces the computation graph, it can fuse and optimize the forward simulation (Materials and Methods, ``Architecture''). Compared to other GPU-accelerated, high-fidelity simulations, this achieves consistent speed-ups of an order of magnitude or more (see \autoref{fig:performance}\textbf{B}). \textsc{Crazyflow} also natively supports full-fidelity vectorized swarm simulations. It scales to thousands of drones in a swarm in thousands of parallel worlds without simplified physics. At the same time, it outperforms other GPU-accelerated simulations such as DiffAero, which only supports swarms with a simplified point-mass model~(PMM), while allowing for a more than tenfold increase in the number of drones (see \autoref{fig:performance}\textbf{C}).

\textsc{Crazyflow}'s computation graph also enables efficient gradient computation during simulation. We compute nine million gradients per second for one million environments, each differentiating through 10 consecutive simulation steps (see \autoref{fig:performance}\textbf{D}). It surpasses the throughput of the state-of-the-art differentiable drone simulator DiffAero~\cite{zhang2025diffaero} by more than 10 times across almost all parallelization levels.

\textsc{Crazyflow} inherits the parallel rendering of depth images from MJX's raycast pipeline (Materials and Methods, ``Sensing''). On an identical test scene, we achieve 350k frames per second (FPS) at a 64x64 ray resolution across 1,024 environments—a fivefold increase in throughput over state-of-the-art DiffAero (see \autoref{fig:performance}\textbf{E}). Beyond depth, \textsc{Crazyflow} supports photorealistic RGB-D rendering via Gaussian splatting, delivering 38k FPS at 100k Gaussians on the GPU compared to 256 for Flightmare. For more benchmarks, scaling on clusters, and statistical analysis, refer to figs. \ref{fig:sup:rotor_vel_benchmark} and \ref{fig:sup:scaling} (Supplementary Text).

\subsection*{Sim-to-Real} \label{sec:results:sim2real}
The primary physics model of \textsc{Crazyflow} is derived from first principles (Materials and Methods, ``First Principles Modeling''). In contrast to typical drone simulators, our model builds on an asymmetric nonlinear motor model that maps individual rotor speed setpoints to the actual speeds. Those speeds are then converted to forces and torques acting on the rigid drone body via thrust curves, with additional detailed aerodynamic and rotor effects. The importance of each part is discussed in tables \ref{tab:sup:sim2real_crazysim} and \ref{tab:sup:sim2real_gpd} (Supplementary Text). All parameters are determined with extensive real-world identification experiments for all three standard Crazyflie (CF) configurations and the recently released brushless model. We achieve high-fidelity simulation by reimplementing the low-level onboard controllers and configurable aerodynamic effects in \textsc{JAX}. We focus our evaluation discussion below on the CF2.1 and CF2.1 Brushless (the two most used Crazyflie configurations), the CF2.1 with the thrust upgrade kit, one custom drone, and the Holybro X500. The CF2.1+ model is available in our repository. The details of the setup and the identification are described in Materials and Methods, ``Mind The Gap: From Real To Sim and Back''.

\begin{table*}[t]
\centering
\caption{\textbf{Sim-to-real performance of different simulators.} \textsc{Crazyflow} reduces the sim-to-real gap by at least $46.0\%$ for the popular Crazyflie (CF) 2.1 platform compared to widely used simulators, such as gym\_pybullet\_drones (GPD) and Learning to Fly (L2F), and even outperforms software-in-the-loop simulators, such as CrazySim, running the latest firmware. For the thrust-upgraded Crazyflie (CFT) and the Crazyflie Brushless (CFB), similar gaps are achieved. The sim-to-real experiments are conducted in attitude control mode using first-principles models, and the gaps are computed as the root mean square position error between the simulation and the real drone over the second of three consecutive laps to reduce the influence of starting and stopping transients. The cycle time of the circle ($\bigcirc$) and Lissajous ($\infty$)} reference is denoted $T$, and the control frequency is 100~Hz. The errors are given as the mean $\pm$ one standard deviation over $N=5$ runs in mm. Since the drone crashes occasionally in CrazySim, we provide a 10 run budget. If not at least 5 runs succeed, we mark the run as failed (---). The full flight test protocol is discussed in the Supplementary Text.
\label{tab:sim2real_sims}
\setlength{\tabcolsep}{4pt}
\vspace{10pt}
\begin{tabular}{l cccc c c c c}
\hline \noalign{\smallskip}
 & \multicolumn{4}{c}{CF 2.1} & & CFT & & CFB \\ 

\cline{2-5} \cline{7-7} \cline{9-9} \noalign{\smallskip}
Trajectory & GPD & CrazySim & L2F & \textsc{Crazyflow} & & \textsc{Crazyflow} & & \textsc{Crazyflow} \\ 
\noalign{\smallskip} \hline \noalign{\smallskip}
$\bigcirc_{\:T=9.0\,\mathrm{s}}$      & $47.1\pm1.0$     & $104.1\pm1.8$ & $87.9\pm1.3$  & $\mathbf{18.3\pm2.6}$ & & $21.9\pm1.3$ & & $30.9\pm1.9$ \\
$\bigcirc_{\:T=3.5\,\mathrm{s}}$      & $122.0\pm2.8$    & ---           & $226.1\pm2.0$ & $\mathbf{60.0\pm1.7}$ & & $60.4\pm2.6$ & & $82.5\pm4.2$ \\
$\infty_{\:T=10.0\,\mathrm{s}}$       & $19.4\pm1.7$     & $66.7\pm1.4$  & $24.7\pm1.5$  & $\mathbf{10.5\pm3.0}$ & & $12.4\pm2.0$ & & $17.2\pm0.6$ \\
$\infty_{\:T=5.5\,\mathrm{s}}$        & $40.1\pm2.0$     & $101.9\pm2.4$ & $52.8\pm2.4$  & $\mathbf{20.6\pm2.7}$ & & $23.9\pm4.3$ & & $33.6\pm3.7$ \\
\noalign{\smallskip} \hline
\end{tabular}
\end{table*}

\begin{table*}[b]
\centering
\caption{\textbf{Sim-to-real comparison of different drones in \textsc{Crazyflow}.} 
Across platforms of different sizes ranging from sub-40~g Crazyflies through a 660~g custom drone to 2.2~kg commercial off-the-shelf drones, \textsc{Crazyflow} provides competitive sim-to-real gaps using our abstracted model. On the Crazyflies, these gaps are as small as a few centimeters and less than half the wheelbase for the larger drones, while the model can be fitted with only a few minutes of flight data. The experiments are conducted the same way as in \autoref{tab:sim2real_sims} and the errors are given as the mean $\pm$ one standard deviation over $N=5$ runs in mm. The full flight test protocol is discussed in the Supplementary Text.}
\label{tab:sim2real_drones}
\vspace{10pt}
\setlength{\tabcolsep}{10pt}
\begin{tabular}{l ccccc}
\hline \noalign{\smallskip}
Trajectory & CF 2.1 & CFT & CFB & Custom & HB X500 \\ 
\noalign{\smallskip} \hline \noalign{\smallskip}
$\bigcirc_{\:T=9.0\,\mathrm{s}}$ & $13.5\pm2.3$  & $16.9\pm1.7$  & $17.9\pm1.7$  & $50.1\pm1.0$  & $58.7\pm4.9$ \\
$\bigcirc_{\:T=3.5\,\mathrm{s}}$ & $19.1\pm2.3$  & $27.4\pm1.1$  & $63.0\pm3.3$  & $87.9\pm2.5$ & $160.3\pm29.4$ \\
$\infty_{\:T=10.0\,\mathrm{s}}$  & $9.5\pm2.9$   & $10.2\pm1.7$  & $10.3\pm1.4$  & $35.0\pm0.9$  & $53.3\pm2.4$ \\
$\infty_{\:T=5.5\,\mathrm{s}}$   & $20.8\pm3.2$  & $21.4\pm5.1$  & $24.8\pm1.5$  & $62.2\pm1.9$  & $76.2\pm3.0$ \\
\noalign{\smallskip} \hline
\end{tabular}
\end{table*}

We validate the fidelity of the full simulation stack by assessing the sim-to-real performance of \textsc{Crazyflow}. We deploy the widely used geometric controller~\cite{mellinger2011controller} in simulation and in the real world to track a circle in the $x$-$y$ plane and a Lissajous in the $x$-$z$ plane at multiple speeds. The sim-to-real gap is computed as the root-mean-squared (RMS) distance between the observed drone positions in the paired simulated and real experiments. For a meaningful comparison, we have adjusted the mass and inertia in all compared simulators to match those of our real drone with its marker deck configuration. Using the first-principles model, \textsc{Crazyflow} achieves a low, centimeter-level gap for the Crazyflie 2.1 quadrotor, see \autoref{tab:sim2real_sims}. The gaps achieved are not only up to $61.2\%$ lower than with gym\_pybullet\_drones, the most popular Crazyflie simulation to date~\cite{mcguire2026bestofrobot} or up to $79.2\%$ lower than with Learning to Fly, but also up to $84.3\%$ lower than with the CrazySim software-in-the-loop simulator running the latest firmware and onboard controls where one of the benchmark runs even crashes in simulation. \textsc{Crazyflow} has a lower sim-to-real gap than all other simulators on the given set of trajectories (Materials and Methods, ``Statistical Analysis''). These improvements stem from rigorous system identification for the high-quality models and architectural design choices of \textsc{Crazyflow} (Supplementary Text). This shows that both our implementation of the low-level control stack and our physics are accurate. In the following sections, we further demonstrate the accuracy of our first-principles model, with and without the low-level control stack, by using it to train deployable RL agents without any domain randomization.

Since obtaining the parameters for the first-principles models is tedious, \textsc{Crazyflow} also supports data-driven physics models for a mid-level control interface commonly used in drone control, explained in full detail in Materials and Methods, ``Abstracted Physics Modeling''. Those abstracted dynamics implicitly fit the low-level control loop, can be fit with just a few minutes of flight data without a motion capture system (Supplementary Text), and achieve a competitive sim-to-real gap compared to the first-principles model, see \autoref{tab:sim2real_sims} and \autoref{tab:sim2real_drones}. In some cases, the sim-to-real gap of the abstracted model is even smaller than that of the first-principles model. 

\begin{figure*}[t]
    \centering
    \includegraphics[width=0.99\textwidth]{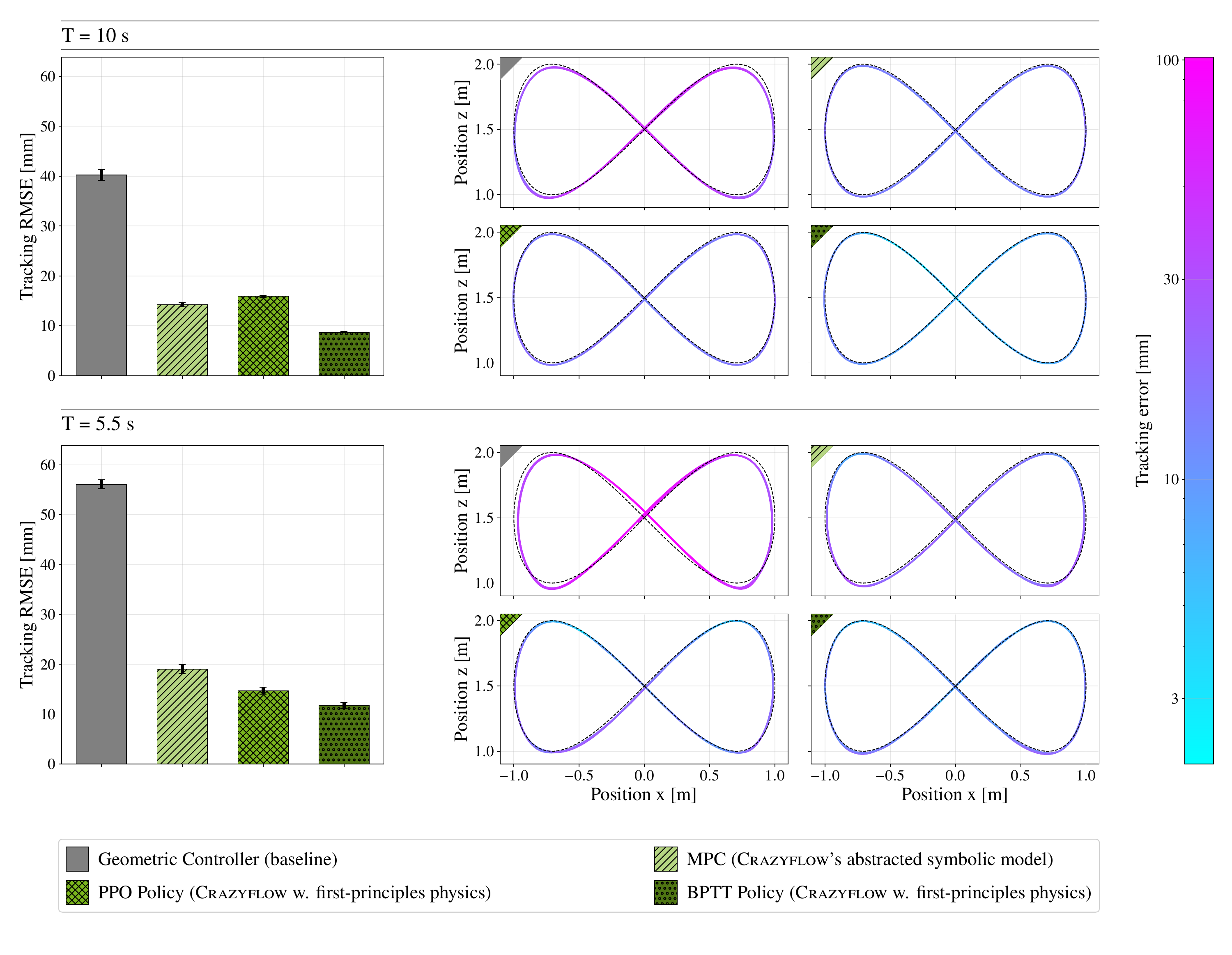}
    \vspace{-2em}
    \caption{\textbf{Real-world tracking performance of learning- and model-based control approaches using \textsc{Crazyflow}}. We compare an NMPC using \textsc{Crazyflow}'s abstracted model, a PPO policy, and a BPTT policy leveraging gradients, both trained in \textsc{Crazyflow} without any domain randomization on the first principles model, against a baseline Geometric Controller on the presented Lissajous tracking task, indicated by the dashed black line. The experiments are conducted on the Crazyflie 2.1 Brushless and reported as mean $\pm$ one standard deviation in mm across $N=5$ runs. The full flight test protocol is discussed in the Supplementary Text.}
    \label{fig:policy}
\end{figure*}

The abstracted model is not restricted to Crazyflie-family nanodrones and also scales to larger drones. We demonstrate this on a custom drone with a 250~mm wheelbase weighing 660~g and the commercial off-the-shelf Holybro X500 with a 500~mm wheelbase weighing 2.2~kg, identified outdoors using global navigation satellite systems for localization. Evaluations with accurate motion capture systems show that both drones had sim-to-real gaps less than half their wheelbase dimensions (\autoref{tab:sim2real_drones}). Along with the \textsc{Crazyflow} simulation, we provide a streamlined system identification pipeline for the data-driven model, which fits the model with under four minutes of flight data. Data collection is not limited to motion capture spaces, as we demonstrate by identifying the Holybro outdoors with global navigation satellite system localization only. This enables quick, low-friction integration of new drones into \textsc{Crazyflow} for rapid adaptation. Although the custom drone is equipped with the popular PX4 flight controller \cite{px4_autopilot}, the pipeline is also applicable to any other controller.

\subsection*{Trajectory Tracking Benchmark}
We verify the efficacy of our simulator on standard trajectory-tracking benchmarks and show how accuracy, speed, and differentiability can be seamlessly leveraged to support algorithm designs for both learning- and control-oriented methods. This capability is demonstrated below by designing attitude-level trajectory-tracking controllers for the Lissajous task from ``Sim-to-Real''. These controllers employ state-of-the-art deep RL and control methods, leveraging first-principles and abstracted models, respectively.

The accuracy of \textsc{Crazyflow} allows us to train real-world-deployable RL agents in seconds without any domain randomization at the limit of the platform's precision. Proximal Policy Optimization (PPO)~\cite{schulman2017proximal}, one of the most popular RL algorithms, achieves a tracking error of $15.9\pm0.2$~mm on the slow trajectory in just 1.61~s of training time and $14.7\pm0.7$~mm on the fast trajectory with 17.58~s of training. Achieving an improvement of up to $73.8\%$ over the geometric controller baseline with a default RL algorithm explicitly highlights the simulator's accurate sim-to-real alignment. Despite training without domain randomization, the policy is also robust to distribution shifts, such as deploying on another drone than the identified one, as well as flying it with a low battery voltage, a center of mass shift, and an 11-gram payload, see fig. \ref{fig:sup:randomization_ablation} (Supplementary Text). By directly utilizing the simulator's analytical gradients via backpropagation through time (BPTT)~\cite{werbos1990bptt}, the training time drops to 1.56~s while further improving tracking performance to $8.7\pm0.1$~mm for the slow trajectory and 12.64~s with $11.8\pm0.6$~mm for the fast trajectory (\autoref{fig:policy}).

Moving beyond learning-based methods, \textsc{Crazyflow} provides symbolic dynamics models implemented in CasADi~\cite{andersson2019casadi}, which can be used directly by optimization-based control methods. We demonstrate this usage with a Nonlinear Model Predictive Controller (NMPC) built with acados~\cite{verschueren2022acados}. Our abstracted models achieve tracking errors of $14.2\pm0.4$~mm and $19.0\pm0.9$~mm for the tasks, respectively, competitive with the learning-based controllers above.

\begin{figure*}[t]
    \centering
    \includegraphics[width=0.99\linewidth]{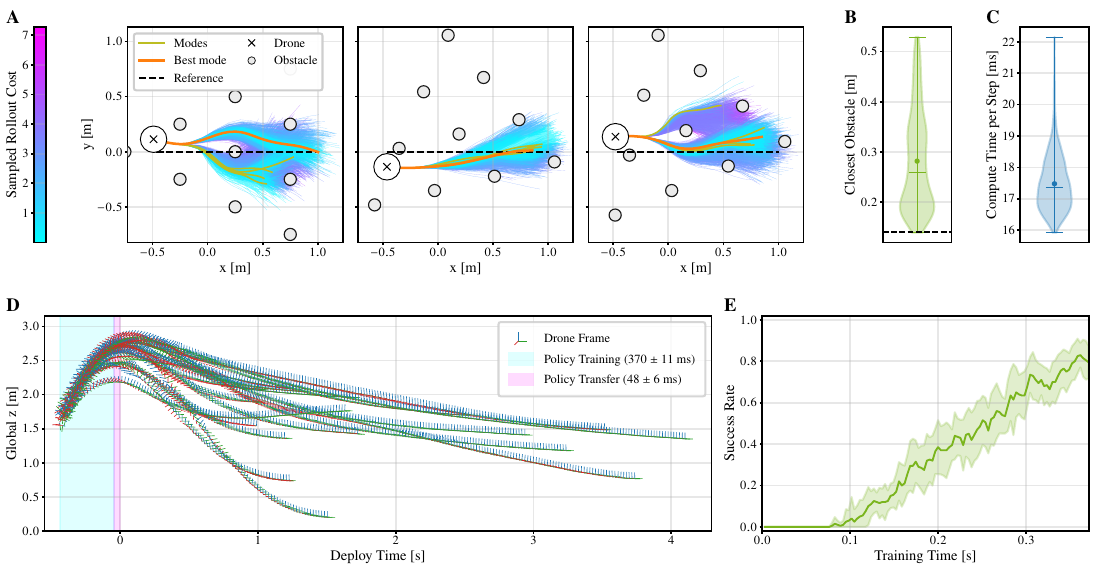}
    \caption{\textbf{High-throughput application examples.} \textsc{Crazyflow}'s speed and accuracy enable real-time sampling-based control and high-performance policy learning. The top row shows real-time control with massively parallel, full model rollouts for MPPI in \textsc{Crazyflow}. We show simulated trajectories for different obstacle scenarios in (\textbf{A}), where each trajectory is colored according to its cost. We also include the distance to the closest obstacle with its limit for safe deployment (\textbf{B}) and the computation time consistent with 50~Hz control (\textbf{C}). \textsc{Crazyflow}'s analytical gradients allow for highly efficient policy training, such that a motor-level stabilization policy can be learned in-flight. The training is initialized with states randomized around the predicted takeover state. (\textbf{D}) shows the 17 successful runs out of 20 real throw experiments. (\textbf{E}) shows the mean and standard deviation of the success rate of the policies trained over ten randomized seeds, evaluated at the 20 observed takeover states.}
    \label{fig:state_experiments}
\end{figure*}

\subsection*{Control in Cluttered Environments}
Beyond basic tracking tasks, \textsc{Crazyflow} facilitates the design of algorithms for challenging tasks such as flying through cluttered environments. These represent interesting constrained optimization problems because they admit multi-modal solutions. We demonstrate this on a high-performance MPPI controller. While traditional model predictive control often relies on linearized or simplified dynamics for online tractability, \textsc{Crazyflow}'s performance allows us to directly use the full non-linear dynamics for MPPI without model simplification. By sampling 500k parallel trajectories at 50~Hz with a 25-step horizon---exceeding half a billion simulation steps per second---the simulator enables real-time, derivative-free optimization in cluttered environments. As a direct result, the controller is highly robust against local minima and solver convergence issues, and can additionally generate multi-modal trajectories. We show this capability by successfully deploying the resulting trajectories on physical hardware for reactive obstacle avoidance (\autoref{fig:state_experiments}\textbf{A}, \textbf{B} and \textbf{C}, movie S3).

\subsection*{High-Speed Training at Motor Level}
The accuracy and computational efficiency of \textsc{Crazyflow} permit the end-to-end training of policies that directly output rotor speed commands, circumventing the conventional need for low-level onboard controllers. By fusing the entire RL loop and simulation into a single JIT-compiled kernel, we eliminate any Python overhead. An RL agent can learn stable position control from raw  rotor commands in 12.16~s, representing a speedup of 14 times over specialized state-of-the-art C++ implementations with 180~s of training time, presented in L2F~\cite{eschmann2024learning_to_fly_in_seconds}. When physically deployed on the same drone model, this policy achieves an improved root mean square position error of $14.47\pm0.60\,\mathrm{cm}$ (5 runs) on the same 5.5~s Lissajous task, which is approximately 15\% lower than the reported accuracy  in~\cite{eschmann2024learning_to_fly_in_seconds}.

Leveraging BPTT further compresses this training to sub-second timescales, demonstrating a shift from ``train-then-deploy'' to ``on-the-fly'' (i.e., in-execution) training paradigms. We show this synergy of gradient-based optimization, speed, and accuracy on a  stabilization task that strictly constrains training times. Upon throwing a physical drone into the air, we start training an agent in \textsc{Crazyflow} with BPTT using the predicted drone state in 500~ms. The policy converges and transfers the parameters to the drone in under half a second, effectively stabilizing the vehicle before it hits the ground. The recovery phase and training progress are shown in \autoref{fig:state_experiments}\textbf{D} and \textbf{E}, where the policy achieves a success rate of 85\% on 20 runs. Despite not randomizing the dynamics, the resulting policies are robust. We prove this through additional throw experiments with intentional domain shifts introduced by removed propeller guards and weaker batteries. Further details on policy training, deployment, and the robustness experiments are provided in the Supplementary Text (figures \ref{fig:sup:throw_randomized_drone} and \ref{fig:sup:throw_no_prediction}, movie S2).

\begin{figure*}[t]
    \centering
    \includegraphics[width=0.99\linewidth]{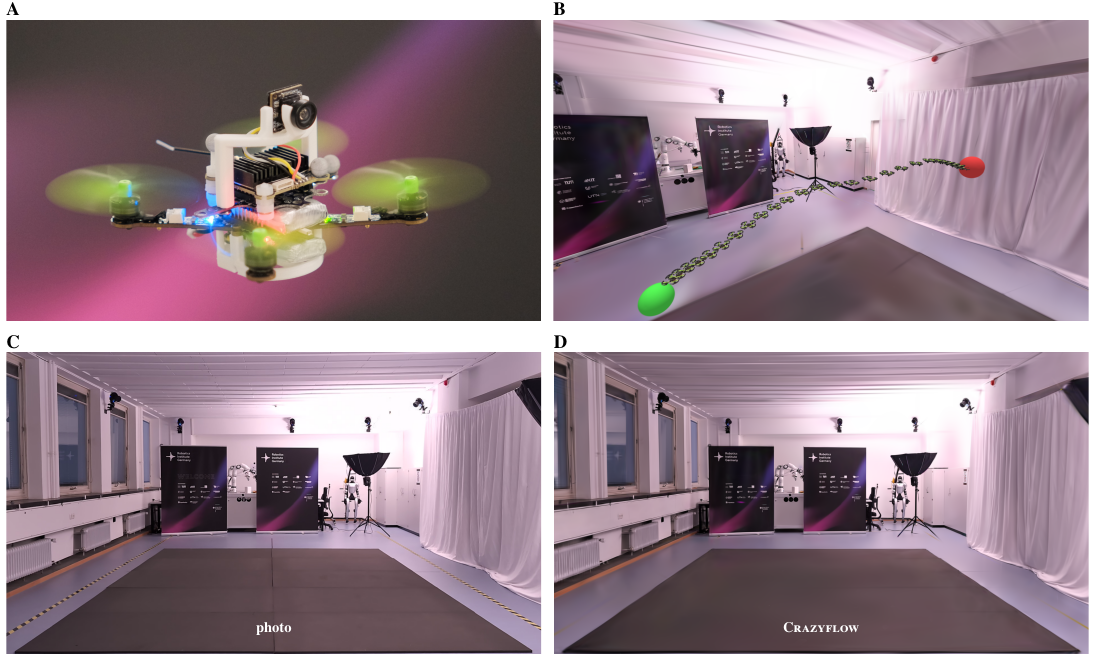}
    \caption{\textbf{End-to-end vision-based flight.} We fly our real-world experiments on a modified CFB with additional battery and onboard camera (\textbf{A}). The task is to navigate from the current position (red) in the flight space to a goal position (green) from camera and onboard sensor inputs only (\textbf{B}). Using a video from our flight space (\textbf{C}), we reconstruct it and use Gaussian splatting to render images in \textsc{Crazyflow} (\textbf{D}).}
    \label{fig:vision}
\end{figure*}

\subsection*{Vision-based Flight}
Beyond state-based policies, we show that \textsc{Crazyflow}'s vision sensors (Materials and Methods, ``Sensing'') allow direct sim-to-real transfer of vision-based policies. In our experiment, we train a policy with a DAgger-like setup~\cite{ross2011dagger} to steer towards a random goal position on images rendered by Gaussian splats of the flight space and inertial measurement unit (IMU) measurements, as shown in \autoref{fig:vision}. Including the reconstruction, it takes less than 1.5 hours to go from no environment model to a trained end-to-end vision-based policy, or about 14 minutes to train the policy alone. We deploy the policy on a CFB modified with an HDZero Eco camera, successfully reaching 10 consecutive goals. Since Crazyflies have limited onboard compute, we stream the camera and sensors to a workstation, which returns attitude commands over the radio link. Full training details are included in the Supplementary Text. A video of the experiment, including the onboard camera and the rendered Gaussian splat, is available in movie S4.

\subsection*{Applications} \label{sec:results:applications}

\textsc{Crazyflow}'s performance and flexibility have already been instrumental in our projects spanning research, competitions, and education.

The architecture scales seamlessly to multi-agent systems, supporting the simulation of swarms with up to thousands of drones. SwarmGPT~\cite{schuck2025swarmgpt} leverages this capability to generate complex aerial swarm performances from natural language descriptions, as seen in \autoref{fig:application}\textbf{E}. The simulator's speed and low sim-to-real gap allow users to iterate quickly on choreography design, verify the generated trajectories, and confidently deploy them in the real world.

Due to the minimal training time of RL agents in \textsc{Crazyflow}, it is well-suited for benchmarking algorithms, where performance can only be evaluated fairly after thorough hyperparameter tuning. We used this to conduct extensive experiments to answer fundamental questions about $\mathrm{SO(3)}$ action representations in RL~\cite{schuck2026primer}. \textsc{Crazyflow} allowed us to ground our investigation in a real robotic problem, ensuring the rigor of the results by executing over 1000 independent hyperparameter and evaluation runs within a highly affordable compute budget.

Beyond research, inspired by our Safe Robot Learning Competition for the 2022 IEEE/RSJ International Conference on Intelligent Robots and Systems~\cite{teetaert2025reproducibility}, \textsc{Crazyflow} powers the Autonomous Drone Racing challenge~\cite{lsy_drone_racing_github}. Students at the Technical University of Munich compete in this high-speed autonomous racing challenge, transitioning from simulation to real-world hardware within a single academic term (see \autoref{fig:application}\textbf{B}). Beyond racing, \textsc{Crazyflow} is extensively leveraged in advanced control lectures~\cite{schoellig2025cfr} to demonstrate the trade-offs between using detailed first-principles models and simplified dynamics models within both control-theoretic and learning-oriented decision-making frameworks. By providing a unified platform for both optimal control and RL, \textsc{Crazyflow} bridges the gap between theoretical algorithm design and practical robotic deployment.

\section*{Discussion}
In this paper, we present \textsc{Crazyflow}, an accurate, GPU-accelerated, differentiable drone simulator in \textsc{JAX}. \textsc{Crazyflow}'s physics are based on detailed dynamics models of quadrotor systems, with a modular structure that supports five commonly used control interfaces, ranging from low-level motor commands to mid-level force-torque and attitude setpoints, to high-level position references. The simulator is highly flexible, allowing reconfigurations and the incorporation of additional aerodynamic effects, domain randomizations, or custom disturbances without adding overhead. By tracing, optimizing, and JIT-compiling the simulator's computation graph in \textsc{JAX}, we achieve state-of-the-art performance at millions of parallel worlds and thousands of drones, and can analytically trace gradients through the full simulation pipeline. To better support control-oriented methods, \textsc{Crazyflow} includes equivalent model implementations in CasADi, an optimization toolbox tailored for model-based control approaches. We provide the physics parameters for the Crazyflie family of nanodrones, parameters for the abstracted model of the Holybro X500 platform, and offer a streamlined identification pipeline to distill abstracted models for any other quadrotor drones.

Our experiments show how \textsc{Crazyflow}'s speed, accuracy, and differentiability come together in increasingly challenging tasks. On a mid-level control interface, high-performing trajectory-tracking PPO and BPTT policies can be trained in under three seconds. Due to accurate physics down to the motor level, policies can be trained directly without domain randomization, using either mid-level attitude or low-level rotor-speed commands. This is further demonstrated by training a motor-level trajectory-tracking policy that achieves state-of-the-art accuracy while outperforming specialized C++ training setups in terms of training time. The synergy of gradient-based optimization, speed, and accuracy unlocks ``on-the-fly'' learning paradigms, exemplified by training an agent to stabilize a drone thrown into the air before it hits the ground.

Beyond supporting learning-based approaches, we show \textsc{Crazyflow} is also directly compatible with control-oriented methods. In the tracking experiments, an NMPC controller with the CasADi models we provided achieved competitive performance compared to the learned policies. Apart from free-space tracking tasks, we further demonstrated that \textsc{Crazyflow} enables sampling-based MPPI algorithms to efficiently solve more challenging, constrained navigation tasks in real time at 50~Hz, using 500k sampling environments in the full-fidelity simulation. 

The flexibility of our simulator in serving the community was further demonstrated through adoption in drone swarm research, hyperparameter optimization for fair RL benchmarking, autonomous drone racing competitions that push algorithms to their limits, and advanced robot decision-making education. The potential of our simulator is not limited to the aforementioned applications. For instance, one could exploit the features of \textsc{Crazyflow} to extend automatic parameter tuning~\cite{berkenkamp2023bayesian} to large-scale gradient-based methods, enable GPU-accelerated swarm behavior optimization for faster, real-time replanning~\cite{hamer2019swarm_trajectories_gpu}, or explore competitive multi-drone racing with online learning of opponent strategies~\cite{spica2018two_player}.

\textsc{Crazyflow} is built on the idea of open-source projects. During development, we contributed to \textsc{SciPy}, the Crazyflie firmware, the array API standard libraries, Gymnasium, and other widely used tools in the community. Our hope is that \textsc{Crazyflow} can become a similarly useful tool for others to contribute to. For instance, the current collection of drone models could grow substantially beyond the current state. In addition, we found the differentiable, parallelized reimplementation of drone controllers vital to our efforts. Including other open-source controller pipelines, such as PX4, BetaFlight, or ArduPilot, would make the simulation's capabilities accessible to a wider range of researchers. With its modular design, \textsc{Crazyflow} provides clear avenues for community contributions and can serve as a hub for further collective scientific advances in aerial robotics research.

\section*{Materials and Methods}
\subsection*{Simulator Design} \label{sec:materials_and_methods:simulator_design}
Below, we give a detailed description of \textsc{Crazyflow}’s architecture and explain our design choices. We also show how this relates to the capabilities we have demonstrated in our applications.

\subsubsection*{Architecture} \label{sec:materials_and_methods:architecture}
\textsc{Crazyflow} is designed with three key capabilities in mind: JIT compilation for efficiency while still maintaining full customizability, support for massively parallel, hardware-accelerated simulation, and differentiability for gradient-based methods. To support these goals, we build on top of \textsc{JAX}'s automatic differentiation and tracing capabilities. \textsc{JAX} follows a strictly functional programming paradigm, centered on stateless functions operating on immutable input data without side effects. 

Following that paradigm, \textsc{Crazyflow} is built around a monolithic simulation data PyTree that holds all buffers required by the simulation (e.g., drone and controller states and drone parameters). We then compose our simulation as a pipeline consisting of functions that take in, transform, and pass on the simulation data. Because the simulation data is compatible with \textsc{JAX}, we can fuse these transforms into a single graph that can leverage an XLA scan operation to efficiently simulate multiple successive steps.

The XLA compiler optimizes this graph for efficient execution on both CPU and GPU hardware. This flexibility allows researchers to achieve both low-latency execution for single agents and massive throughput for parallel environments. While general-purpose \textsc{JAX} simulators such as MJX must resolve contact constraints using iterative methods, \textsc{Crazyflow} numerically integrates the drone's equations of motion directly. Removing the solver loop improves execution speed and enhances numerical stability of gradients during backpropagation.

This concept of a modular pipeline of functions is illustrated in \autoref{fig:overview}. We depict modules as cubes to visualize the three data dimensions on which they operate: parallel worlds, drones per world, and data axes. The modularity extends internally to components like the low-level controllers, allowing us to support control inputs at different levels of abstraction. Similarly, modular dynamics allow us to seamlessly swap in identified models from our real-to-sim pipeline. The computation graph captures all active modules, fusing computations for JIT compilation while allowing differentiation across the entire stack.

While \textsc{Crazyflow} is implemented in \textsc{JAX}, we recognize that this may not be a one-size-fits-all choice for all researchers. Our core physics models and controller implementations are thus built on Python's array API standard~\cite{meurer2023arrayapi}, which enables backend-agnostic code. In practice, this means our modules natively support not only \textsc{JAX}, but also \textsc{PyTorch}, \textsc{NumPy}, \textsc{CuPy}, and any other standard-compliant framework. As a result, researchers can use our core physics and controller models with, e.g., \textsc{PyTorch}, and the underlying operations are dynamically dispatched to \textsc{PyTorch} instead of \textsc{JAX}. This design aligns with a major current effort across scientific libraries such as \textsc{SciPy} and \textsc{scikit-learn} to support framework-agnostic code.

These architectural choices directly enable the performance and sim-to-real capabilities demonstrated throughout our experiments. 

\subsubsection*{Customizing the Simulation} \label{sec:materials_and_methods:customization}
Breaking the simulation step into transforms is a powerful tool for customization. On initialization, we dynamically select only the required functions for the current configuration to build the simulation pipeline, reducing complexity and memory usage. At the same time, this design allows researchers to customize the simulation by adding or removing their own transforms to the controllers, physics, and reset logic. 

\textsc{Crazyflow}'s pipelines provide these domain-specific customizations without the performance penalties typical of modular software. Because \textsc{JAX} performs kernel fusion during compilation, user-defined modifications such as custom disturbances or randomizing environmental parameters are integrated directly into the optimized simulation step. This also allows us to eliminate any overhead from the array API-compatible core physics and controller modules.

We use this modularity in our experimental setups. Our tasks use different levels of cascaded control, including direct rotor velocity and attitude. \textsc{Crazyflow} dynamically assembles its simulation step using the required components for each task. Furthermore, in the drone racing challenge, we exploit the compiler's kernel fusion to integrate custom dynamic disturbances and noise sources directly into the simulation step, maintaining zero-overhead execution despite the added environmental complexity.

\subsubsection*{Collision Checking}
Beyond high-fidelity free-flight physics, many learning environments require collision detection. We achieve these capabilities by leveraging MJX, a \textsc{JAX}-compatible reimplementation of the MuJoCo physics engine. The simulation environment is defined using MuJoCo's XML environment descriptions, loaded into MJX, and selectively synchronizes its state with MJX to selectively utilize MuJoCo’s collision-checking algorithms.

\subsubsection*{Sensing}
In addition to collision checking, vision-based approaches also require simulated camera sensors. We cover this on two levels. First, we can likewise use MJX for efficient ray casting and distance measurements within the \textsc{JAX} computation graph. Since both \textsc{Crazyflow} and MJX are JIT-compatible, the entire sensing and physics pipeline is compiled into a single call.

While MJX gives us rudimentary depth sensing, many applications in vision-based flight require sensors that closely align with the real world. We support this with sensors that render RGB-D images directly from a trained reconstruction of the flight space through Gaussian splatting. Splatting has several desirable properties, e.g. it can reconstruct scenes from videos instead of modeling them by hand, trades compute for realism through the number of Gaussians in a scene, parallelizes efficiently by rendering the same splat from different viewpoints, and makes vision sensors differentiable. Its implementation, however, requires dynamic shapes inside the computation graph, which is impossible in JAX. Instead, we implement the core logic in NVIDIA Warp~\cite{macklin2022warp}, and use foreign function interfaces to make it available in JAX~\cite{splax2026github}. Our splatting kernels are batched by default, allow for efficient relocation of Gaussians inside the Warp kernels, and let us render camera sensors across worlds and drones effectively.

\subsection*{Mind The Gap: From Real To Sim and Back} \label{sec:materials_and_methods:real2sim}
While the computational throughput provided by \textsc{JAX} is critical for modern robotic learning, the utility of a simulator is ultimately limited by the \textit{sim-to-real gap}, i.e., the discrepancy between the simulated and the real-world dynamics. If this gap is too large, the policies learned in simulation fail to transfer reliably to physical platforms. To bridge this gap, \textsc{Crazyflow} provides a high-fidelity modeling backend based on real-world data. We offer multiple models, ranging from first-principles physics-based to data-driven models. Additionally, thanks to the flexible simulation pipeline, users can easily add their own physics backend that better suits their needs. 

\subsubsection*{First Principles Modeling} \label{sec:materials_and_methods:real2sim_first_principles}
The default modeling approach in \textsc{Crazyflow} characterizes the system through analytical, first-principles dynamics. While many existing simulators simplify the platform by assuming idealized actuator responses, \textsc{Crazyflow} incorporates high-fidelity details down to the motor level. As our results in ``Sim-to-Real'' indicate, capturing these physical effects is essential for highly accurate modeling (Supplementary Text). We represent the drone as a rigid body with six degrees of freedom, where the sum of forces $\mathbf{f}_\Sigma$ and torques ${}^{\mathcal{B}}\mathbf{t}_\Sigma$ acting on the body results in

\begin{align}
    \dot{\mathbf{p}} &= \mathbf{v}, \\
    \dot{\mathbf{q}} &= \frac{1}{2} \mathbf{q} \otimes \begin{bmatrix}{}^{\mathcal{B}}\bm{\omega} \\ 0 \end{bmatrix}, \\
    m\dot{\mathbf{v}} &= \mathbf{f}_\Sigma, \\
    \mathbf{J} \, {}^{\mathcal{B}}\dot{\bm{\omega}} &= {}^{\mathcal{B}}\mathbf{t}_\Sigma - {}^{\mathcal{B}}\bm{\omega}\times\mathbf{J} \, {}^{\mathcal{B}} \bm{\omega},
    \label{eq:first_principles_model}
\end{align}
where $\mathbf{p}$ is the position, $\mathbf{v}$ is the velocity, $\mathbf{q}$ is the orientation as a scalar-last quaternion, and ${}^{\mathcal{B}} \bm{\omega}$ is the angular velocity in body frame with the parameters $m$ and $\mathbf{J}$ as the mass and inertia matrix of the drone, respectively. The quaternion product is denoted by $\otimes$ and states in the body frame by~${}^\mathcal{B}(\cdot)$.

The main force acting on the body, in addition to the gravitational force $\mathbf{f}_\mathrm{g}=m\mathbf{g}$ with the gravitational constant $\mathbf{g}$, is the thrust generated by the four propellers mounted at the corners of the drone frame, generating a force ${}^{\mathcal{B}}\mathbf{f}_\mathrm{t} = \mathbf{e}_\mathrm{z} \sum_i^4 f_{\mathrm{m},i}$ and a torque ${}^{\mathcal{B}}\mathbf{t}_\mathrm{t} = l \, \mathrm{diag} (1,1,0) \, \mathbf{M} \, \mathbf{f}_\mathrm{m}$, 
where $\mathbf{e}_\mathrm{z}$ is the unit vector in $z$ direction, $l$ is the distance of the motors to the body axes, $\mathrm{diag}(\cdot)$ is the diagonal matrix, and $\mathbf{M}$ is the drone specific mixing matrix, which defines the motor placement and spin direction relative to the drone's body frame. The thrust produced by each propeller $f_{\mathrm{m},i}$ is a function of its speed $\Omega_i$, modeled by a second-order polynomial. Besides the axial forces, the spinning propellers induce torques. Similarly to the force, each motor produces an aerodynamic counter torque ${}^{\mathcal{B}}\mathbf{t}_\mathrm{d} = \mathrm{diag}(0,0,1) \, \mathbf{M} \, \mathbf{t}_{\mathrm{m}}$ due to rotational drag with the surrounding air, where $t_{\mathrm{m},i}$ is also modeled as a second-order polynomial of $\Omega_i$. Additionally, gyroscopic and reaction torques are defined as 
${}^{\mathcal{B}}\mathbf{t}_\mathrm{g} = J_\mathrm{p} \left({}^{\mathcal{B}}\bm{\omega} \times \mathbf{e}_\mathrm{z}\right) \mathbf{e}_\mathrm{z}^\top \, \mathbf{M} \, \bm{\Omega}$ 
and 
${}^{\mathcal{B}}\mathbf{t}_\mathrm{r} = J_\mathrm{p} \, \mathbf{e}_\mathrm{z}\, \mathbf{e}_\mathrm{z}^\top \, \mathbf{M} \, \dot{\bm{\Omega}}$, 
respectively, 
where $J_\mathrm{p}$ is the combined inertia of one propeller and its motor.

Another important force to consider is the drone's aerodynamic drag. We model this effect as a linear term in the body frame ${}^{\mathcal{B}}\mathbf{f}_\mathrm{a}=\mathbf{C}_\mathrm{a} {}^{\mathcal{B}} \mathbf{v}$, where $\mathbf{C}_\mathrm{a}$ contains the drag coefficients in matrix form. In low speed regimes, more complex models drag models, as used in \cite{folk2023rotorpy}, do not improve the accuracy of the model \cite{torrente2021drag}. 
The total force and torque acting on the drone results in $\mathbf{f}_\Sigma = \mathbf{f}_\mathrm{g} + \mathbf{R}\, {}^{\mathcal{B}}\mathbf{f}_\mathrm{t} + \mathbf{R}\, {}^{\mathcal{B}}\mathbf{f}_\mathrm{a}$ and ${}^{\mathcal{B}}\mathbf{t}_\Sigma = {}^{\mathcal{B}}\mathbf{t}_\mathrm{t} + {}^{\mathcal{B}}\mathbf{t}_\mathrm{d} + {}^{\mathcal{B}}\mathbf{t}_\mathrm{g} + {}^{\mathcal{B}}\mathbf{t}_\mathrm{r}$, respectively, where $\mathbf{R}$ is the rotation from body to world frame.

The fidelity of the rigid body model is fundamentally tied to the rotor angular velocities, which are themselves governed by complex underlying dynamics. From an idealized physics perspective, these rotor dynamics may be characterized as $J_\mathrm{p}\dot{\Omega}_i = \tau_{\mathrm{m},i} - c_\mathrm{v}\Omega - c_\mathrm{d}\Omega|\Omega|$, where $\tau_{\mathrm{m},i}$ is the motor torque command, following \cite{graefe2026brushless}, but incorporating an additional drag term. However, this conventional approach faces two significant challenges in practice. First, standard quadrotor motor controllers typically only apply positive torque during acceleration, causing the rotor dynamics to differ significantly during acceleration and deceleration. Second, the steady-state equilibrium of such a model is highly sensitive to the identified parameters, hindering generalizability across different physical instances of the same platform. To address these complexities, \textsc{Crazyflow} adapts the physical model to
\begin{equation}
    \dot{\bm{\Omega}} = \begin{cases}
        \hat{c}_\mathrm{v} (\bm{\Omega}_{\mathrm{cmd}} - \bm{\Omega}) + \hat{c}_\mathrm{d} (\bm{\Omega}_{\mathrm{cmd}}^2 - \bm{\Omega}^2) & \forall \,\bm{\Omega}_{\mathrm{cmd}} \geq \bm{\Omega} ,\\
        \check{c}_\mathrm{v} (\bm{\Omega}_{\mathrm{cmd}} - \bm{\Omega}) + \check{c}_\mathrm{d} (\bm{\Omega}_{\mathrm{cmd}}^2 - \bm{\Omega}^2) & \forall \,\bm{\Omega}_{\mathrm{cmd}} < \bm{\Omega},
    \end{cases} \label{eq:rotor_adapted}
\end{equation}
where $\hat{c}_\mathrm{v}$, $\check{c}_\mathrm{v}$, $\hat{c}_\mathrm{d}$, and $\check{c}_\mathrm{d}$ are the viscous damping and drag parameters of the rotor dynamics, respectively. This model can capture different dynamics for acceleration and deceleration, while also allowing for pure quadratic braking behavior induced by the propeller's drag. The state and input to the dynamical system result in 
$\mathbf{x} = \begin{bmatrix} \mathbf{p}^\top , \mathbf{q}^\top , \mathbf{v}^\top , {}^{\mathcal{B}}\bm{\omega}^\top , \bm{\Omega}^\top \end{bmatrix}^\top$ and $\mathbf{u} = \bm{\Omega}_{\mathrm{cmd}}$, respectively.

While the presented model operates at the lowest level of control, which typically requires update rates of at least 500~Hz, this level of control authority is rarely necessary for most applications. In the majority of cases, even in high-performance control, low-level controllers are employed to abstract the dynamics~\cite{sun2022Comparison_FMPC_NMPC_INDI}. As shown in \autoref{fig:overview}, \textsc{Crazyflow} provides five control interfaces. Operating at increasing levels of abstraction, it includes rotor velocity, collective thrust and body torques (torque interface), collective thrust and body rates (rates interface), collective thrust and attitude (attitude interface), and position setpoints. We recommend control frequencies of 500~Hz, 250~Hz, and 50~Hz for the torque, rates, and attitude interfaces, respectively. The full control stack for the Mellinger control type \cite{mellinger2011controller} on the Crazyflie platform has been reimplemented using the array API standard to enable full differentiation and JIT compilation via \textsc{JAX}. Due to the aforementioned modularity, the controllers can also be easily swapped and adapted. 

The parameters for the first-principles model are obtained in a wide set of experiments. While some parameters, such as the mass $m$, are easy to obtain from direct measurements or well approximated from CAD data, as in the case of the propeller's inertia $J_\mathrm{p}$, other parameters are more tedious to identify. For fitting the thrust and torque curve (i.e., the mapping from rotor speed to thrust and torque), each drone is mounted on an ATI Nano43 load cell to collect static and dynamic thrust data across multiple motors and batteries. The inertia of the rigid body $\mathbf{J}$, arguably the most difficult parameter to obtain, is collected directly from flight data, as introduced in \cite{Eschmann_2024}. Similarly, the aerodynamic drag coefficients $\mathbf{C}_\mathrm{a}$ are obtained from aggressive flight data.

\subsubsection*{Abstracted Physics Modeling}
As established in the previous section, first-principles models require the implementation of low-level controllers and can involve extensive parameter identification, which may introduce unnecessary complexity for many research tasks. Furthermore, practitioners frequently use high-level interfaces for planning \cite{schuck2025swarmgpt} or mid-level attitude interfaces for control-focused tasks \cite{sun2022Comparison_FMPC_NMPC_INDI}. We therefore additionally provide an abstracted dynamics model that hides the inner control loop within stable dynamics, interfacing at the attitude level. Based on the assumption of a stable and well-tuned low-level control loop, the drone's attitude, represented by Euler angles $\bm{\Psi}$, is characterized as a second-order linear system. While Euler angles have downsides compared to quaternions, such as nonlinearities and gimbal lock, these effects are negligible near the upright orientation, making them a practical choice for many drone applications.

Another abstraction is made on the rotor level. Instead of modeling rotor velocities, we abstract their effect to a single collective thrust state $f_{\Sigma}$, which is subject to first-order delay. The input to the model consists of the desired attitude and thrust, which results in $\mathbf{u}_\mathrm{abstracted} = \begin{bmatrix} \bm{\Psi}^\top_\mathrm{cmd}, f_{\Sigma,\mathrm{cmd}} \end{bmatrix}^\top$. Given the attitude and collective thrust, we assume the drone to be a point-mass subject to the same linear drag term as in the first-principles model. The state space model results in 
\begin{align}
    \dot{\mathbf{p}} &= \mathbf{v}, \\
    m\dot{\mathbf{v}} &= \mathbf{f}_\mathrm{g} + \mathbf{R}\, \mathbf{e}_z c_\mathrm{f}f_{\Sigma} + \mathbf{R}\,{}^{\mathcal{B}}\mathbf{f}_\mathrm{a}, \label{eq:3_simplified_acceleration} \\
    \dot{f}_\Sigma &= c_\tau (f_{\Sigma,\mathrm{cmd}} - f_\Sigma), \label{eq:3_simplified_force} \\
    \ddot{\bm{\Psi}} &= \bm{c}_{\bm{\Psi},1} \bm{\Psi} + \bm{c}_{\bm{\Psi},2} \dot{\bm{\Psi}} + \bm{c}_{\bm{\Psi},3} \bm{\Psi}_\mathrm{cmd} \label{eq:3_simplified_rpy},
\end{align}
where $\mathbf{R}$ is the rotation from body to world frame, $c_\tau$ and $c_\mathrm{f}$ are the thrust time and scaling coefficient, respectively, and $\bm{c}_{\bm{\Psi},1}$, $\bm{c}_{\bm{\Psi},1}$, and $\bm{c}_{\bm{\Psi},3}$ are the coefficients for the rotational dynamics. The state of the abstracted model spans $\mathbf{x}_\mathrm{abstracted} = \begin{bmatrix} \mathbf{p}^\top, \bm{\Psi}^\top, \mathbf{v}^\top, \dot{\bm{\Psi}}^\top, f_{\Sigma} \end{bmatrix}^\top$. For further simplification, we can remove the thrust dynamics and the drag term, albeit at the cost of reduced fidelity.

Building on its simplified structure, the parameters for this model can be obtained from just a few minutes of flight data. Empirical tests show that a combination of Lissajous (5:6), melon, and modified Rhodonea trajectories at multiple speeds over a total of 206~s yields a dataset rich enough to robustly fit the model. For validation, different types, i.e., Lissajous (1:2), spiral, and Clélie trajectories, are tracked at different speeds, totaling 227~s. We also provide a geometric controller based on \cite{mellinger2011controller} to track those references, with three gain presets for small, medium, and large drones, which only requires the drone's mass as an additional input.

\subsection*{Statistical Analysis}
To check the significance of our benchmarking results (\autoref{fig:performance}), we use one-sided Welch $t$-tests on log timings. For a general statement on improved performance across all scenarios, we use an intersection-union test (IUT). Similarly, we assess the sim-to-real differences in \autoref{tab:sim2real_sims} using one-sided paired Student's $t$-tests on log RMS distances, with five paired observations per trajectory. We combine the trajectory-wise tests using an IUT to assess whether \textsc{Crazyflow} is closer to the real drone on every evaluated trajectory. The global $p$-values are $p_\mathrm{IUT}<0.001$ against \textsc{gym\_pybullet\_drones}, Learning to Fly, and \textsc{CrazySim}. The tests cover four trajectories, except for \textsc{CrazySim}, where the fast circle is excluded because every attempt crashed. Further details of the analyses are included in the Supplementary Text.

\bibliography{references}

\clearpage
\section*{Acknowledgments}
We thank the \textsc{SciPy} team for their support with the rewrite of the \texttt{scipy.spatial.transform} module, the Bitcraze team for the discussions and quick incorporation of improvements into the Crazyflie firmware, and Wolfgang Hönig for sharing the hardware required for us to identify the physical parameters of the CF2.1 Brushless.

\paragraph*{Funding:} This work was funded by the Robotics Institute Germany under BMBF grant 16ME0997K, and by the Humboldt Professorship for Robotics and Artificial Intelligence.

\paragraph*{Author contributions:} M. Schuck had the idea for \textsc{Crazyflow}, designed the architecture, implemented the simulation, contributed required changes to \textsc{SciPy}, wrote parts of the RL training scripts, performed hardware experiments, and wrote the manuscript. M. Rath developed the models, implemented the differentiable controllers, wrote the system identification pipeline, performed extensive model identification on the Crazyflies, ran hardware experiments, contributed to the Crazyflie drone firmware, and wrote the manuscript. Y. Hua implemented large parts of the RL and BPTT training and performed hardware experiments. A. Goudar prepared the hardware experimental results on the self-built drone. S. Zhou provided close supervision on the project, advised on technical issues, and wrote the manuscript. A. P. Schoellig was the main advisor on the project, helped prepare the manuscript, and organized the funding.

\paragraph*{Competing interests:}
There are no competing interests to declare.

\paragraph*{Data and materials availability:}
All other data needed to evaluate the conclusions in the paper are present in the paper or the Supplementary Materials. The data for this study have been deposited in Dryad (DOI: 10.5061/dryad.ksn02v7m2).
\textsc{Crazyflow} can be installed via PyPI. The source code is available on GitHub under \url{https://github.com/learnsyslab/crazyflow}. The full code to reproduce the experiments is public under
\url{https://github.com/learnsyslab/crazyflow_experiments}. Our changes to \textsc{SciPy} have been incorporated into the main project and are available from version 1.17 onwards. Our improvements and measurements of the Crazyflies have also been incorporated into the official firmware.

\section*{Supplementary Text}

\subsection*{Simulator Comparison}
To give a complete picture, \autoref{tab:sup:simulator-capabilities} provides a detailed comparison of capabilities against state-of-the-art simulators used as baselines in the main text. In summary, as compared to native Python simulators, we support both CPUs and GPUs, with strong scaling performance for single drones and swarms. Compared to C++-based simulators, we inherit the flexibility of Python, show that we can match or outperform them in terms of throughput, and generalize to cluster-size deployments (see ``Benchmark Ablations''). We also demonstrate stronger rendering performance. \textsc{Crazyflow} models the physics more accurately than other simulators, as evident from our sim-to-real benchmarks in the main text, the Materials and Methods, ``First Principles Modeling'', and from the following sections on ``Simulator Physics Ablation'' and ``Sim-to-Real Ablations on Rotor-level Simulators.''  In the following sections, we provide further justifications on the dynamics modeling, additional ablations on sim-to-real transfer, as well as speed and scalability benchmarking.

\begin{table*}[t]
  \centering
  \small
  \setlength{\tabcolsep}{3pt}
  \renewcommand{\arraystretch}{0.75}
  \caption{Simulator capabilities. \cyes{} supported, \cpart{} implemented with limitations, \cno{} not implemented.}
  \label{tab:sup:simulator-capabilities}

  \resizebox{\textwidth}{!}{%
  \begin{tabular}{lccccccc}
    \hline \noalign{\vspace{0.5em}}
     & \textsc{Crazyflow}
     & L2F~\cite{eschmann2024learning_to_fly_in_seconds}
     & Flightmare~\cite{song2021flightmare}
     & GPD~\cite{panerati2021learning}
     & DiffAero~\cite{zhang2025diffaero}
     & Aerial Gym~\cite{kulkarni2025aerialgym}
     & CrazySim~\cite{llanes2024crazysim} \\
     & {\tiny Python (JAX)}
     & {\tiny C++ bindings}
     & {\tiny C++ bindings}
     & {\tiny Python (NumPy)}
     & {\tiny Python (torch)}
     & {\tiny Python (torch)}
     & {\tiny Python (NumPy)} \\
    \hline \noalign{\vspace{0.5em}}

    \multicolumn{8}{l}{\textit{Command interfaces}} \\
    Position                   & \cyes & \cno  & \cno   & \cyes  & \cno   & \cyes  & \cyes \\
    Attitude / thrust          & \cyes & \cno  & \cno   & \cpart & \cno   & \cyes  & \cyes \\
    Body rates / thrust        & \cyes & \cno  & \cpart & \cpart & \cyes  & \cyes  & \cyes \\
    Rotor level                & \cyes & \cyes & \cyes  & \cyes  & \cno   & \cyes  & \cno \\
    \hline \noalign{\vspace{0.5em}}

    \multicolumn{8}{l}{\textit{Actuator effects}} \\
    Rotor speed dynamics       & \cyes & \cyes & \cyes & \cno & \cno & \cyes & \cyes \\
    Nonlinear rotor dynamics   & \cyes & \cno  & \cno  & \cno & \cno & \cno  & \cno \\
    Asymmetric spin-up / -down & \cyes & \cyes & \cno  & \cno & \cno & \cyes & \cyes \\
    Rotor reaction torque      & \cyes & \cno  & \cno  & \cno & \cno & \cno  & \cno \\
    Gyroscopic effect & \cyes & \cno  & \cno  & \cno & \cno & \cno  & \cyes \\
    \hline \noalign{\vspace{0.5em}}

    \multicolumn{8}{l}{\textit{Aerodynamic effects}} \\
    Drag                       & \cyes & \cno & \cno & \cyes & \cyes & \cyes & \cyes \\
    Ground effect              & \cyes & \cno & \cno & \cyes & \cno  & \cno  & \cyes \\
    Downwash                   & \cyes & \cno & \cno & \cyes & \cno  & \cno  & \cpart \\
    \hline \noalign{\vspace{0.5em}}

    \multicolumn{8}{l}{\textit{Domain randomization}} \\
    Arbitrary parameters       & \cyes & \cyes & \cpart & \cpart & \cpart & \cpart & \cpart \\
    Per-rotor resolution       & \cyes & \cyes & \cno   & \cno   & \cno   & \cyes  & \cpart \\
    Per-world, at runtime      & \cyes & \cyes & \cpart & \cpart & \cyes  & \cpart & \cno \\
    \hline \noalign{\vspace{0.5em}}

    \multicolumn{8}{l}{\textit{Simulation and compute}} \\
    Independent physics rate   & \cyes & \cno   & \cpart & \cyes  & \cyes & \cyes & \cpart \\
    GPU physics                & \cyes & \cpart & \cno   & \cno   & \cyes & \cyes & \cno \\
    Distributed GPUs           & \cyes & \cno   & \cno   & \cno   & \cno  & \cno  & \cno \\
    Differentiable dynamics    & \cyes & \cno   & \cno   & \cno   & \cyes & \cno  & \cno \\
    Vectorized environments    & \cyes & \cyes  & \cyes  & \cpart & \cyes & \cyes & \cno \\
    Swarms                     & \cyes & \cno   & \cno   & \cyes  & \cpart & \cno & \cyes \\
    Double precision           & \cyes & \cno   & \cno   & \cyes  & \cno & \cno & \cyes \\
    \hline \noalign{\vspace{0.5em}}

    \multicolumn{8}{l}{\textit{Firmware}} \\
    Software in the loop$^{a}$ & \cpart & \cno & \cno & \cpart & \cno & \cno & \cyes \\
    \hline \noalign{\vspace{0.5em}}

    \multicolumn{8}{l}{\textit{Sensing}} \\
    RGB rendering              & \cyes & \cno & \cpart & \cyes & \cno & \cyes & \cpart \\
    Depth rendering            & \cyes & \cno & \cpart & \cyes & \cyes & \cyes & \cno \\
    Batched RGB / depth        & \cyes & \cno & \cpart & \cno & \cyes & \cyes & \cno \\
    \hline
  \end{tabular}%
  }

  \vspace{0.4em}
  {\footnotesize
  $^{a}$While we internally test our controllers against the real firmware
  to make sure both match, we do not consider this full SITL. We opt for a
  JAX reimplementation since directly using the firmware is at odds with our
  primary goals: parallelization, differentiability, JIT fusing, etc.
  }
\end{table*}

\subsubsection*{Simulator Physics Ablation} \label{sec:materials_and_methods:real2sim_ablations}
As demonstrated above, \textsc{Crazyflow} has a lower sim-to-real gap than the compared simulators in \autoref{tab:sim2real_sims}, \textsc{gym\_pybullet\_drones} (GPD)~\cite{panerati2021learning}, \textsc{CrazySim}~\cite{llanes2024crazysim}, and Learning to Fly (L2F)~\cite{eschmann2024learning_to_fly_in_seconds}. In this subsection, we discuss which components cause these differences.

\begin{figure*}[ht!]
    \centering
    \includegraphics[width=\linewidth]{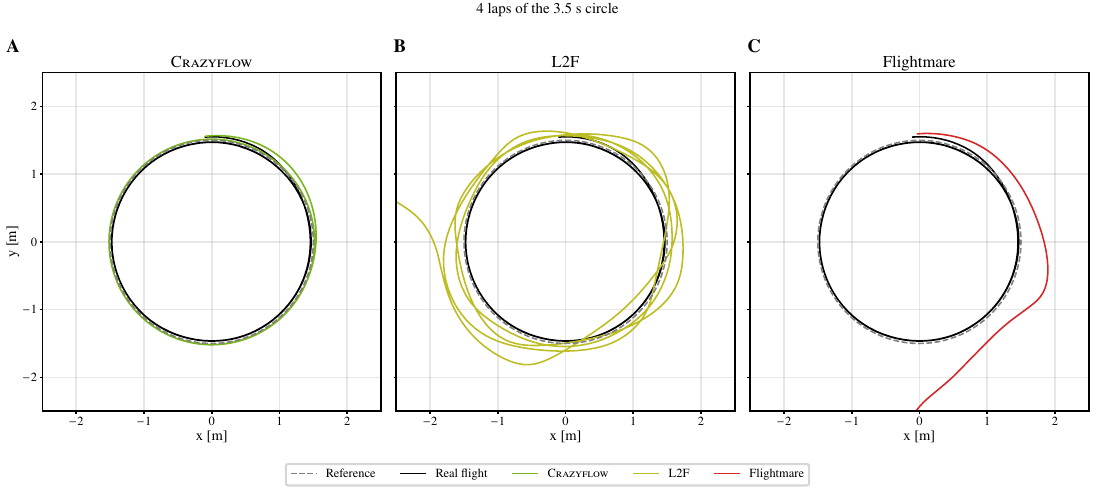}
    \caption{Comparison of a fast, real trajectory on the CF2.1 with \textsc{Crazyflow} (\textbf{A}), L2F (\textbf{B}) and Flightmare (\textbf{C}). Missing reaction-torque terms in L2F and Flightmare lead to instability due to the large motor time constants of the CF2.1, even though the real drone manages to complete the trajectory.}
    \label{fig:sup:rotor_vel_sims}
\end{figure*}

The first difference is the treatment of low-level actuator dynamics. Existing simulators either reduce the rotors to first-order linear systems, as in L2F and CrazySim, or drop the rotor state by assuming any commanded thrust is realized instantaneously, as in GPD. This assumption introduces modeling errors with direct consequences for sim-to-real transfer and can be especially restrictive for training motor-level policies. \textsc{Crazyflow} instead models motors as nonlinear first-order systems and further differentiates the asymmetry in the motor dynamics when spinning up and spinning down. A second effect \textsc{Crazyflow} explicitly models and existing simulators commonly omit is the counter torque a rotor exerts on the airframe while it accelerates. For platforms with relatively large rotor time constants, such as the Crazyflie~2.1, omitting this effect can produce substantial modeling errors and unstable behavior during aggressive, high-speed maneuvers. Third, \textsc{Crazyflow} includes gyroscopic effects induced by the rotors. To quantify the individual contributions to the sim-to-real performance, we ablate each effect in \autoref{tab:sup:sim2real_crazysim}. Starting from CrazySim as a representative baseline, we add our modeled effects and measure the resulting sim-to-real error in leave-one-out ablations. Excluding the rotor-acceleration reaction torque increases the sim-to-real error and results in instability on the more challenging trajectories. As we further demonstrate in \autoref{fig:sup:rotor_vel_sims}, this term is also absent from other commonly used simulators, including L2F and Flightmare, leading to unrealistic rotational behavior under these operating conditions. While removing the nonlinear, asymmetric spin-up and spin-down model and gyroscopic effects has only a minor influence on the closed-loop trajectories,  these richer sets of effects allow us to more faithfully model the measured rotor dynamics.

\begin{table*}[h!]
\centering
\caption{\textbf{Sim-to-real ablation study results for \textsc{CrazySim}.} We compare both CrazySim Gazebo-based and MuJoCo-based physics backends on the Crazyflie 2.1. The default backend for CrazySim is Gazebo-based. It should be noted that the CrazySim's MuJoCo-based backend simulation partially adopted our identified drone parameters, which contributed to the observed differences in columns two and three of the table. To determine which modeling effects distinguish \textsc{Crazyflow} from CrazySim, we modify CrazySim to match \textsc{Crazyflow} in four respects and perform a leave-one-component-out ablation: (1) properly synchronizing clock time, (2) adding the reaction torque of the accelerating rotors, which the stock backend omits, (3) replacing its symmetric first-order linear rotor model by our identified asymmetric nonlinear rotor dynamics, and (4) correcting the sign of the propeller gyroscopic torque, which the stock backend applies reversed.  The values are the RMS distances between the simulation and the real Crazyflie 2.1, all with the outer position loop at $100\,\mathrm{Hz}$ and the attitude loop at $500\,\mathrm{Hz}$. The errors are given as the mean $\pm$ one standard deviation over 5 runs in mm, the best value per row in bold. Since the CrazySim firmware-in-the-loop chain is not deterministic and crashes stochastically, each simulation run is paired with one real flight. Runs that do not take off or crash are discarded and repeated with a budget of 10 attempts per trajectory; entries with ``---'' denote that fewer runs than real flights completed within the budget. We use the abbreviations $\bigcirc$ for the circle and $\infty$ for the Lissajous trajectories, where the subscript denotes the loop time $T$.
}
\label{tab:sup:sim2real_crazysim}
\setlength{\tabcolsep}{2pt}
\vspace{10pt}
\begin{tabular}{l cc @{}p{6pt}@{} ccccc}
\hline \noalign{\smallskip}
 & \multicolumn{2}{c}{CrazySim Model}
 & & \multicolumn{5}{c}{CrazySim with Our Effects (1)--(4) Added} \\
\cline{2-3} \cline{5-9} \noalign{\smallskip}
Traject. & Gazebo & MuJoCo & & all & w/o (1) & w/o (2) & w/o (3) & w/o (4) \\
\noalign{\smallskip} \hline \noalign{\smallskip}
$\bigcirc_{\:T=9.0\,\mathrm{s}}$
 & $104.1\pm1.8$ & --- & & $15.8\pm2.2$ & $34.3\pm1.4$ & --- & $\mathbf{15.4\pm1.7}$ & $15.9\pm2.0$ \\
\noalign{\smallskip}
$\bigcirc_{\:T=3.5\,\mathrm{s}}$
 & --- & --- & & $\mathbf{125.2\pm1.3}$ & $251.0\pm10.2$ & --- & $125.3\pm2.9$ & $125.7\pm3.5$ \\
\noalign{\smallskip}
$\infty_{\:T=10.0\,\mathrm{s}}$
 & $66.7\pm1.4$ & $22.1\pm2.4$ & & $\mathbf{9.3\pm3.1}$ & $22.7\pm1.7$ & $9.4\pm3.0$ & $9.4\pm3.0$ & $9.4\pm3.0$ \\
\noalign{\smallskip}
$\infty_{\:T=5.5\,\mathrm{s}}$
 & $101.9\pm2.4$ & $34.9\pm3.1$ & & $21.9\pm3.1$ & $37.5\pm1.3$ & $21.9\pm3.2$ & $22.1\pm3.2$ & $\mathbf{21.7\pm3.1}$ \\
\noalign{\smallskip} \hline
\end{tabular}
\end{table*}

\begin{table*}[h!]
    \centering
    \caption{\textbf{Sim-to-real ablation study results for gym\_pybullet\_drones.} We compare the default gym\_pybullet\_drones model on the Crazyflie 2.1, without aerodynamic drag and with its optional drag model (\texttt{PYB\_DRAG}), against the same simulator with \textsc{Crazyflow}'s identified drag matrix added. 
    The values are the RMS distances between the simulation and the real Crazyflie 2.1, all with the outer position loop at $100\,\mathrm{Hz}$ and the attitude loop at $500\,\mathrm{Hz}$. The errors are given as the mean $\pm$ one standard deviation over 5 runs in mm, the best value per row in bold. We use the abbreviations $\bigcirc$ for the circle and $\infty$ for the Lissajous trajectories, where the subscript denotes the loop time $T$. 
    }
    \label{tab:sup:sim2real_gpd}
    \setlength{\tabcolsep}{4pt}
    \vspace{10pt}
    \begin{tabular}{l cc @{}p{6pt}@{} c}
    \hline \noalign{\smallskip}
     & \multicolumn{2}{c}{Default GPD Model}
     & & \multicolumn{1}{c}{GPD with Our Effects Added} \\
    \cline{2-3} \cline{5-5} \noalign{\smallskip}
    Trajectory & no drag & \texttt{PYB\_DRAG} & & identified drag \\
    \noalign{\smallskip} \hline \noalign{\smallskip}
    $\bigcirc_{\:T=9.0\,\mathrm{s}}$
     & $47.1\pm1.0$ & $32.0\pm1.4$ & & $\mathbf{13.7\pm3.6}$ \\
    \noalign{\smallskip}
    $\bigcirc_{\:T=3.5\,\mathrm{s}}$
     & $122.0\pm2.8$ & $83.4\pm2.2$ & & $\mathbf{41.2\pm1.4}$ \\
    \noalign{\smallskip}
    $\infty_{\:T=10.0\,\mathrm{s}}$
     & $19.4\pm1.7$ & $14.0\pm2.3$ & & $\mathbf{10.6\pm2.9}$ \\
    \noalign{\smallskip}
    $\infty_{\:T=5.5\,\mathrm{s}}$
     & $40.1\pm2.0$ & $31.4\pm2.4$ & & $\mathbf{25.9\pm2.8}$ \\
    \noalign{\smallskip} \hline
    \end{tabular}
\end{table*}

\textsc{Crazyflow} also models aerodynamic effects, including drag, using parameters identified directly from data collected on the physical platforms. In contrast, several existing simulators employ drag models  reported in earlier literature (e.g., \cite{forster2015crazyflie} as used in GPD~\cite{panerati2021learning}). As shown in \autoref{tab:sup:sim2real_gpd}, such models are often not sufficiently accurate. In this ablation, replacing the original drag model in GPD with our identified drag model, while leaving the remainder of the simulation pipeline unchanged, significantly reduces the sim-to-real error. This result demonstrates that identifying the relevant aerodynamic effects through real-world testing and system identification is another important contributor to the improved simulation accuracy.

The remaining differences are not physical, but architectural. We implement the onboard controllers in the same JAX framework as the dynamics, so that every control interface in \autoref{fig:overview} can be fused into a single computation graph. CrazySim is the natural reference point here, as it runs the actual firmware in the loop. The ablation study in \autoref{tab:sup:sim2real_crazysim} (columns four and five) further shows that the case with our implemented controller is more faithful than the software-in-the-loop simulation. In our investigation, we trace the poorer accuracy back to clock synchronization. In particular, CrazySim decouples the simulator clock from its software-in-the-loop clock, producing measurements and commands with inconsistent timings if the physics simulation runs with a real-time factor $<1.0$. In our experiments, we measure $0.99$ for the Gazebo backend and $0.91$ for the MuJoCo backend. Note that while this is machine dependent, we have tested this on an Intel(R) Core(TM) Ultra 9 285K. Larger deviations from $1.0$ result in less precise simulated estimators. The deviation with the MuJoCo backend is sufficiently large to affect the sim-to-real gap significantly. We demonstrate this by setting the firmware clock to be in sync with the simulator clock, which improves the sim-to-real results in \autoref{tab:sup:sim2real_crazysim}, column 4. \textsc{Crazyflow}'s architecture prevents this error class by directly fusing a JAX-based version of the onboard controllers into the simulation at synchronized steps. Overall, with \textsc{Crazyflow}, we can directly compile the controller (at the selected interface level) all the way down to individual rotor dynamics into a single computation graph; this design enables both reliable and large-scale parallel simulations.

We also want to point out that the default Gazebo backend has two bugs that contribute to its subpar performance: A typo in the config sets the blade flapping term for one of the propellers ten times larger than for the others, which causes an asymmetric moment, and the thrust chain (PWM to rotor speed mapping, rotor speed to thrust mapping, and rotor speed limit) in the Gazebo plugin is hard-coded to the values for the CFT.

\subsubsection*{Sim-to-Real Ablations on Rotor-level Simulators} We can also compare the sim-to-real error of the rotor-level simulators, L2F and Flightmare, for the CF2.1. Flightmare does not have a set of parameters for this drone and hence uses our identified parameters. All three simulations use our reimplementation of the firmware's state controller, and receive the desired rotor velocities or forces to fly the trajectories from the sim-to-real section. The results are shown in \autoref{fig:sup:rotor_vel_sims}. It should be emphasized that this trajectory is intentionally close to the limits of the platform in the real world.

In Flightmare, the drone crashes after 2.2 s on the fast circle trajectory. Similarly, L2F crashes after 13.8s. Both simulators omit a reaction-torque term arising from accelerating the rotors, which is necessary to stabilize the yaw control of the drone (see ${}^{\mathcal{B}}\mathbf{t}_\mathrm{r}$ in Materials and Methods, ``First Principles Modeling''). Flightmare focuses on racing drones with extremely small motor time constants, where the reaction-torque effect can be neglected without adverse impact; however, using the same modeling approach for platforms with larger motor time constants will lead to instability as observed in \autoref{fig:sup:rotor_vel_sims}. L2F is more stable than Flightmare, as the simulator uses its own identified motor time constant that is smaller in magnitude than ours. Even though it is more stable, as we run the simulation for a longer duration, the simulated trajectories diverge even though the physical drone can successfully complete the 4-lap tracking task. \textsc{Crazyflow} maintains its small sim-to-real error and displays a stable trajectory. Thus, for drones such as the CF2.1, where motor time constant can be non-negligible, incorporating motor reaction torque is important for reliably simulating the drones' behavior.

\subsubsection*{Sim-to-Real Statistical Analysis} \label{sec:materials_and_methods:real2sim_statistics}
\begin{table*}[hb!]
\centering
\caption{\textbf{Paired sim-to-real comparison of \textsc{Crazyflow} against the other simulators.} For every trajectory we give the geometric mean ratio (GMR) of the RMS distances between simulation and the real Crazyflie 2.1 of \textsc{Crazyflow} over those of each simulator, with its 95\,\% confidence interval (CI). A ratio below one means that \textsc{Crazyflow} is closer to the real drone, e.g., 0.5 halves the error. Each simulation run is paired with one real flight. The GMR is the exponential of the mean paired log difference of the RMS distances over the 5 real flights, and the CI is the two-sided 95\,\% Student's $t$ interval for this mean with four degrees of freedom, exponentiated to obtain a ratio interval. All simulators run their default model, for CrazySim the Gazebo-based backend, with the outer position loop at $100\,\mathrm{Hz}$ and the attitude loop at $500\,\mathrm{Hz}$. Entries with ``---'' denote that fewer CrazySim runs than real flights completed within the budget of 10 attempts. We use the abbreviations $\bigcirc$ for the circle and $\infty$ for the Lissajous trajectories, where the subscript denotes the loop time $T$.}
\label{tab:sup:sim2real_gmr}
\setlength{\tabcolsep}{4pt}
\vspace{10pt}
\begin{tabular}{l@{\hspace{12pt}}cc @{}p{24pt}@{} cc @{}p{24pt}@{} cc}
\hline \noalign{\smallskip}
 & \multicolumn{2}{c}{GPD} & & \multicolumn{2}{c}{L2F} & & \multicolumn{2}{c}{\textsc{CrazySim}} \\
\cline{2-3} \cline{5-6} \cline{8-9} \noalign{\smallskip}
Trajectory & GMR & 95\,\% CI & & GMR & 95\,\% CI & & GMR & 95\,\% CI \\
\noalign{\smallskip} \hline \noalign{\smallskip}
$\bigcirc_{\:T=9.0\,\mathrm{s}}$
 & $0.38$ & $[0.32, 0.46]$ & & $0.21$ & $[0.17, 0.25]$ & & $0.17$ & $[0.14, 0.22]$ \\
\noalign{\smallskip}
$\bigcirc_{\:T=3.5\,\mathrm{s}}$
 & $0.49$ & $[0.48, 0.50]$ & & $0.27$ & $[0.26, 0.27]$ & & --- & --- \\
\noalign{\smallskip}
$\infty_{\:T=10.0\,\mathrm{s}}$
 & $0.52$ & $[0.41, 0.66]$ & & $0.41$ & $[0.29, 0.57]$ & & $0.15$ & $[0.11, 0.22]$ \\
\noalign{\smallskip}
$\infty_{\:T=5.5\,\mathrm{s}}$
 & $0.51$ & $[0.46, 0.57]$ & & $0.39$ & $[0.34, 0.44]$ & & $0.20$ & $[0.17, 0.23]$ \\
\noalign{\smallskip} \hline
\end{tabular}
\end{table*}

For each trajectory $j\in\{1,2,3,4\}$, we fly the same physical drone five times (see Flight Test Protocol) and roll out every simulator's first principles model once per real flight in a freshly started simulator instance from the same initial state using the same controller. The tracking mismatch of a rollout is the RMS distance between the simulated and the real position over the flight. Because the $i$-th rollout of every simulator is compared against the $i$-th real flight, the samples of two simulators are paired by real flight, where $i\in\{1,2,3,4,5\}$. Separate simulation runs capture run-to-run variability in the non-deterministic simulator (\textsc{CrazySim}). \autoref{tab:sim2real_sims} reports the mean and standard deviation of the RMS distance per simulator and trajectory.

The RMS distance is strictly positive with a spread growing with its magnitude, and the simulators differ by factors. We therefore test on the log scale, using the paired differences $d_i = \log r^{A}_i - \log r^{B}_i = \log (r^{A}_i / r^{B}_i)$ between \textsc{Crazyflow} ($A$) and a baseline ($B$), which turn the test into a statement about the ratio of the RMS distances. We assume that, within each trajectory, the paired log differences are independent across flights and approximately normally distributed. For each trajectory $j$ we run a one-sided paired $t$-test with
\begin{equation*}
  H_{0,j}\colon \mu_j \geq 0 \qquad H_{1,j}\colon \mu_j < 0,
\end{equation*}
where $\mu_j$ is the expected paired log difference, so that $\mu_j < 0$ means the geometric mean RMS distance of \textsc{Crazyflow} is smaller than that of the baseline on trajectory $j$. We only call \textsc{Crazyflow} closer to the real drone if it is closer on \emph{every} trajectory. This global claim is the intersection $H_1 = \bigcap_j H_{1,j}$, tested against $H_0 = \bigcup_j H_{0,j}$ with an intersection-union test (IUT)~\cite{berger1982}. The IUT rejects $H_0$ at level $\alpha$ if and only if every $H_{0,j}$ is rejected at level $\alpha$, so its $p$-value is $p_{\mathrm{IUT}} = \max_j p_j$ and no multiplicity correction is needed.

The IUT yields $p_{\mathrm{IUT}} < 0.001$ against gym\_pybullet\_drones, Learning to Fly, and CrazySim each, where the test covers only three trajectories because CrazySim crashed on every attempt at the fast circle. For $\alpha = 0.05$ we therefore reject $H_0$ for all baselines and conclude that \textsc{Crazyflow} has a lower RMS distance to the real drone than the other simulators on the presented set of trajectories. Additionally, in \autoref{tab:sup:sim2real_gmr}, the individual geometric mean ratios with confidence intervals between \textsc{Crazyflow} and the three compared simulators are presented.

\subsection*{Experimental Setup}
\subsubsection*{System Identification}
Building on its simplified structure, the parameters for the abstracted model can be obtained from just a few minutes of flight data. Empirical tests show that a combination of Lissajous (5:6), melon, and modified Rhodonea trajectories at multiple speeds over a total of 227~s yields a dataset rich enough to robustly fit the model. For validation, different types, i.e., modified Lissajous (1:2), spiral, and modified Clélie trajectories, are tracked at different speeds, totaling 270~s. We also provide a geometric controller based on \cite{mellinger2011controller} to track those references, with three gain presets for the Crazyflies, the custom drone, and the Holybro drone, which only requires the drone's mass as an additional input.

The drone's pose can either be read from a motion capture system, as done for the Crazyflies and the custom drone, or from a pose estimate based on other forms of localization, such as global navigation satellite systems, as for the Holybro. Note that the data collection in principle supports any other form of pose estimation, e.g., based on ultra-wideband localization. Together with the controller's commands, the data is preprocessed by removing outliers and filtering with a state variable filter, which removes high-frequency measurement noise and generates exact derivatives up to second order. For more details on the exact implementation, we refer the reader to our accompanying repository~\cite{crazyflow_github}.

The system identification process is decoupled into two independent stages: identifying the rotational and the translational dynamics. For the rotational dynamics, assuming closed-loop stability allows us to roll out the model over the entire trajectory for a given set of parameters. We then use this rollout to compute residuals between the observed and predicted orientations and minimize the fitting error using analytical gradients from \textsc{JAX} via Trust Region Reflective optimization~\cite{2020SciPy-NMeth}. Conversely, an open-loop rollout of the translational dynamics would rapidly diverge. To circumvent this issue, we restrict the rollout to the stable thrust state. Taking the norm of \eqref{eq:3_simplified_acceleration} eliminates the rotational dependencies, allowing the acceleration residuals to be computed directly from the known velocity, thrust, and gravity. As with the rotational dynamics, the analytical gradients of these residuals are subsequently used to optimize the translational parameters.

\subsubsection*{Benchmark Protocol}
All benchmarks are running on the same machine, with a Ryzen 9 7950X CPU, an RTX 4090 GPU and 128~GB of DDR5 memory. We configure all simulations to use the same frequencies for their control and physics steps, respectively. We benchmark across the number of parallel environments, swarm sizes where available, and devices, using powers of~4 as step size for the parallel worlds, powers of 2 for the drone swarm sizes, and the CPU/GPU. Render benchmarks use a resolution of 64x64 pixels. All simulators get a budget of 10 minutes to complete 50 repeats of 50 steps per configuration. Where applicable, we warm up JIT caches, e.g. for torch in DiffAero and JAX in \textsc{Crazyflow}.

\subsubsection*{Flight Test Protocol}
Although all simulators were compared using the same deployment data, we conducted the sim-to-real experiments (\autoref{tab:sim2real_sims} and \autoref{tab:sim2real_drones}) in a fixed cyclic order to reduce bias across trajectory types. We repeated trajectories 1, 2, 3, and 4 in sequence until the low-battery indicator activated. After replacing the battery, we resumed the sequence. For the trajectory-tracking benchmark (\autoref{fig:policy}), we used a similar protocol. We evaluated each of the two trajectories separately, cycling through the policies to reduce bias across policies. Whenever the low-battery indicator activated, we replaced the battery and resumed the policy sequence. We followed the same protocol in the ablation studies in the Supplementary Text wherever possible. For ablations requiring physical modifications to the drone, such as adding mass or removing the propeller guard, runs for each configuration were performed consecutively.

\subsection*{Benchmark Ablations}
\subsubsection*{Rotor-level Simulation}

\begin{figure*}[t]
    \centering
    \includegraphics[width=\linewidth]{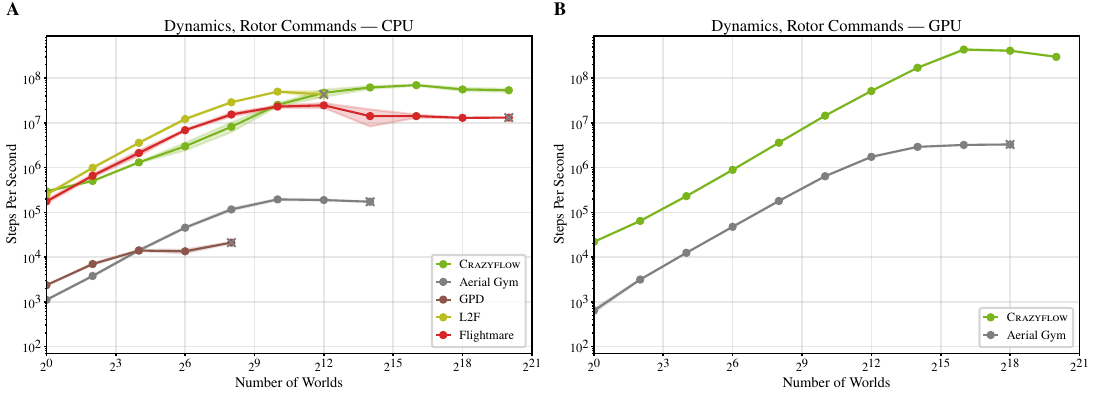}
    \caption{Performance benchmark of simulations supporting rotor-level control on the CPU (\textbf{A}) and the GPU (\textbf{B}). We follow the same protocol as the attitude-level benchmarks in the paper.}
    \label{fig:sup:rotor_vel_benchmark}
\end{figure*}

The benchmarks in the main text focus on simulators that are readily accessible from Python and can be adapted to different research workflows, particularly in machine learning, where Python is the standard ecosystem. However, this excludes other popular simulators such as L2F and Flightmare. These are not straightforward to setup for comparisons because their primary use-case is compiled C++/CUDA. Here we compare against their provided Python bindings.

Neither simulator exposes an attitude command interface, so the attitude-level comparison of the main text does not carry over. \autoref{fig:sup:rotor_vel_benchmark} instead evaluates every simulator that supports rotor-level control. For pure dynamics computations on the CPU, \textsc{Crazyflow} achieves about a 10x speedup over Flightmare for massively parallel workloads on the same device. We hypothesize that this is because Flightmare uses an array-of-structs memory layout that leads to more cache misses, whereas \textsc{Crazyflow} uses a struct-of-arrays approach that lends itself better to parallelization with, e.g., AVX vector instructions. On smaller numbers of parallel environments, L2F is competitive. However, its bindings are capped at 4k simulations and do not include GPU support for further scaling, whereas \textsc{Crazyflow} scales beyond 100M steps per second.

\subsubsection*{Cluster Scaling} 
\begin{figure*}[ht!]
    \centering
    \includegraphics[width=\linewidth]{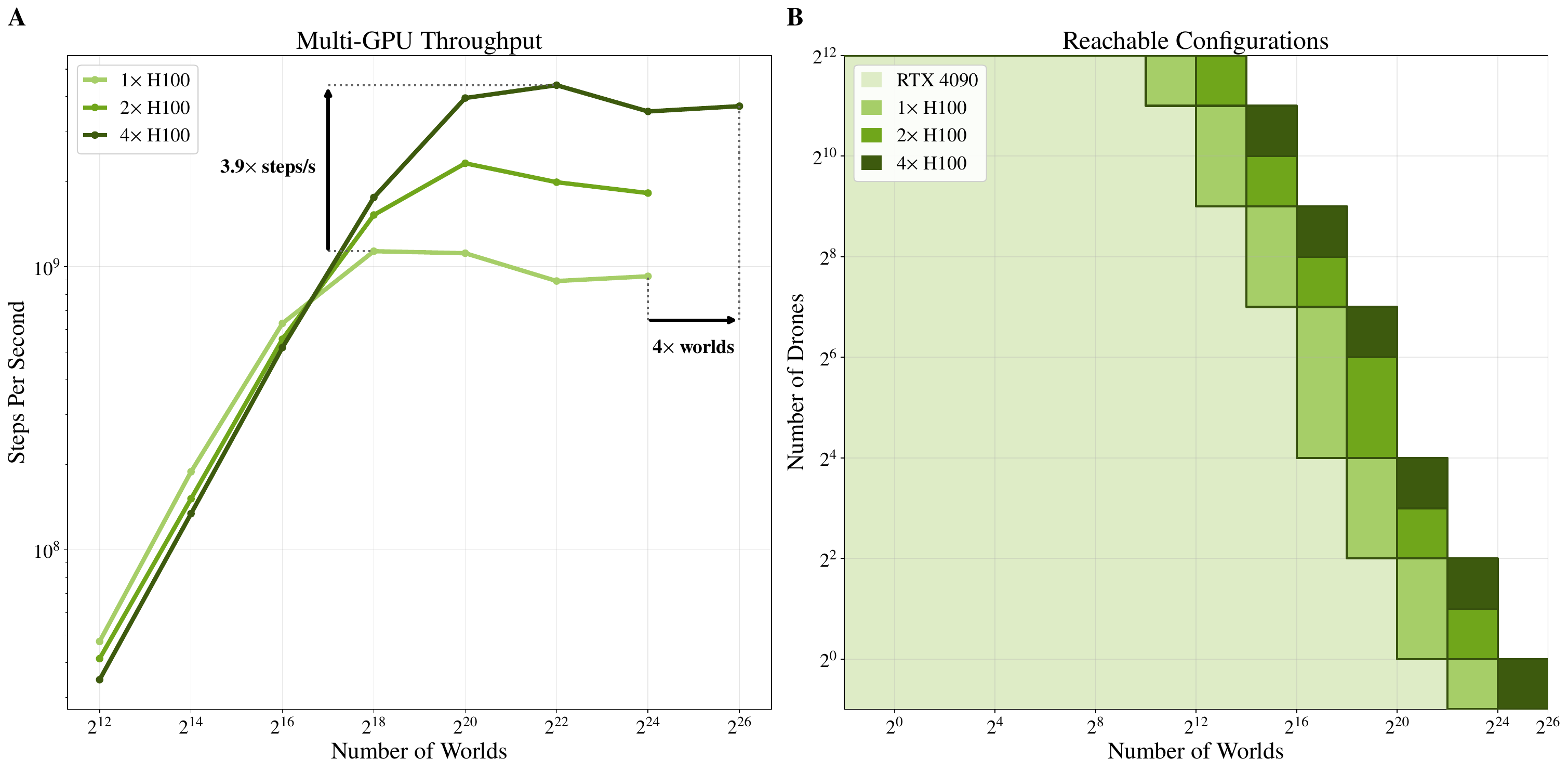}
    \caption{Scaling benchmarks on a 4xH100 HGX node. Sharding \textsc{Crazyflow} across GPUs allows near-linear throughput scaling and reaching more parallel environments (\textbf{A}). It also expands the swarm capabilities, with up to 128 drones in one million parallel simulations (\textbf{B}).}
    \label{fig:sup:scaling}
\end{figure*}

Scaling has become a dominant theme in ML research. While current drone research is mostly conducted on single machines, future algorithms are likely to exploit increasing compute resources. \textsc{Crazyflow} supports this by allowing the simulation to be sharded across GPUs on clusters. We demonstrate this on a 4xH100 HGX node. \autoref{fig:sup:scaling} shows that sharding allows us to scale linearly the number of drones with the number of devices beyond 67 million environments, and that throughput similarly increases by a factor of four to 4.4 billion steps per second. Sharding also further expands the range of reachable configurations, e.g. by reaching 128 drones at 1M parallel environments. Note that speedups are constrained to cluster use-cases with massive parallelization. In regimes with smaller environment counts, timings are dominated by the launch overhead and the clock frequencies of the GPUs, and communication overhead slows us down slightly.

\subsubsection*{Statistical Significance of Results} We test the statistical significance of our benchmark results point-wise on the three speed-up claims against DiffAero, and on the general statement that \textsc{Crazyflow} outperforms other simulators computationally in \autoref{fig:performance}. Since simulation speeds differ by factors and sample variances differ between implementations, we use one-sided Welch t-tests on the logarithm of the steps per second. At 1024 worlds, the throughput exceeds DiffAero by at least 25$\times$ on the CPU, 33$\times$ on the GPU and 18$\times$ on the gradient computation (one-sided log Welch t-tests, Holm-Bonferroni corrected, all $p<0.01$). Similar to the sim-to-real analysis, we can also perform an IUT, yielding that \textsc{Crazyflow} is faster at every world- and drone count in \autoref{fig:performance}\textbf{A}-\textbf{D} with $p<0.001$.

We considered confounders that typically bias the timings in benchmarking. All benchmarks run on the same machine in a freshly spawned process and with no background processes. Timings are preceded by a warm-up period. Slower drifts through thermal throttling are not controlled, but since benchmarks on thousands of worlds run on the order of minutes, we expect devices to reach a thermal steady-state, and expect the effect to change the results by a few percent at most.

\subsection*{Generalization Ablations}
Above, we stated that, besides the vision experiments, all policies are trained without domain randomization. To demonstrate that the policies are not brittle to distribution shifts, we have conducted further ablations.

\subsubsection*{Trajectory Tracking Policies}
\begin{figure*}
    \centering
    \includegraphics[width=1.0\linewidth]{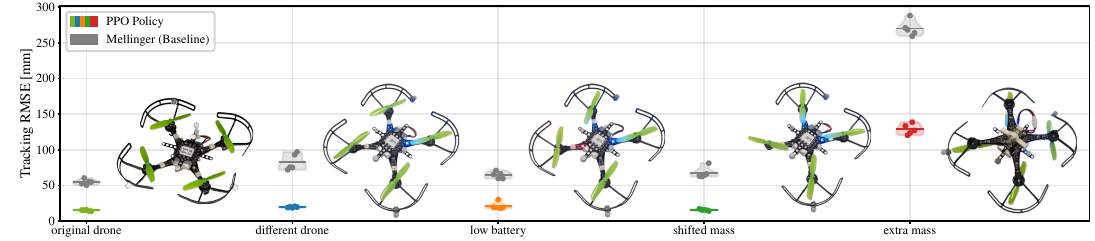}
    \caption{Effects of different drone configurations on a $5.5\,\mathrm{s}$ Lissajous tracking task achieved with the identical PPO policy presented in the full paper. The RMS tracking error and a photo of each corresponding drone is shown for each configuration. They consist of a different drone and the same drone with a low battery (indicated by the red LED), with one propeller guard removed, and with an $11\,\mathrm{g}$ ($\sim25\%$ of the drone's mass) added weight.}
    \label{fig:sup:randomization_ablation}
\end{figure*}

We have tested the robustness of our PPO trajectory-tracking policies, specifically how the tracking error changes for different drone randomizations in the absence of dynamics domain randomization during training. In particular, we tested a different drone and additionally evaluated flights with a low battery, with a changed mass distribution by removing a propeller guard, and with an added mass of $11\,\mathrm{g}$ ($\sim25\%$ of the drone's mass). All of these modifications induce genuine distribution shifts in the dynamics. The modified drones are shown in \autoref{fig:sup:randomization_ablation}, which also shows that the tracking error remains low in most cases. While the average RMSE increases for the strongest disturbance, i.e., the added mass, compared to the nominal case, the policy remains stable and still significantly outperforms the baseline controller (Mellinger). Notably, the Mellinger controller also degrades substantially under these modifications, indicating that the altered dynamics make the tracking task itself considerably more challenging rather than exposing a failure specific to our policy. In the low-battery case, we further observe that our battery compensation --- contributed to the upstream open-source Crazyflie firmware --- works well. This experiment demonstrates that the trajectory tracking policies trained without domain randomization are not brittle.

\subsubsection*{Throw Policies}
\begin{figure*}[h!]
    \centering
    \includegraphics[width=1.0\linewidth]{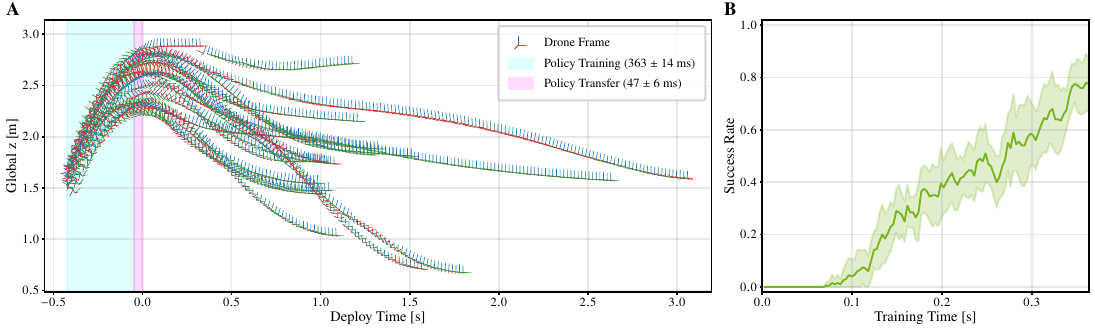}
    \caption{\textbf{Throw policies under drone variations}. (\textbf{A}) shows the 16 successful runs out of 20 real throw experiments using a drone with one propeller guard removed and a lighter battery. (\textbf{B}) shows the mean $\pm$ one standard deviation across ten seeds of the policy's success rate in simulation over training time, evaluated at all 20 observed takeover states. Refer to \autoref{fig:state_experiments} for more details.}
    \label{fig:sup:throw_randomized_drone}
\end{figure*}

\begin{figure*}[h!]
    \centering
    \includegraphics[width=1.0\linewidth]{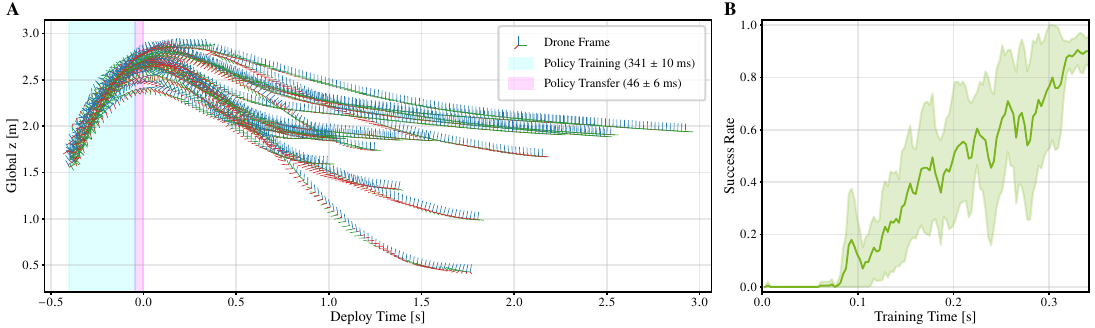}
    \caption{\textbf{Throw policies without takeover state prediction}. (\textbf{A}) shows the 17 successful runs out of 20 real throw experiments without takeover state prediction, i.e., the policy is trained on random throws. (\textbf{B}) shows the mean $\pm$ one standard deviation across ten seeds of the policy's success rate in simulation over training time, evaluated at all 20 observed takeover states. Refer to \autoref{fig:state_experiments} for more details.}
    \label{fig:sup:throw_no_prediction}
\end{figure*}

The BPTT policies trained in the throw experiments are robust against distribution shifts. We ran additional experiments where we remove one propeller guard and use a lighter battery. This introduces a systematic difference between the real and the simulated drone's mass and inertia. On the same throw experiment, this yields a success rate of 80\% across 20 trials, see \autoref{fig:sup:throw_randomized_drone}. This is only one failure more than above and demonstrates that our policy is robust to distribution shifts. Our throw demonstration also trains a stable recovery policy \textbf{without} the predicted takeover state. Using the same setup including the timing constraints, we use randomly initialized start states instead of perturbations around the takeover state. The resulting policies, as shown in \autoref{fig:sup:throw_no_prediction}, stabilize the drone in 17 out of 20 cases, equal to the regular setup. This proves we train a full feedback recovery policy instead of overfitting to an online optimized trajectory for a single throw.

\subsection*{Evaluation Algorithms}
All learning algorithms and their respective training loops are implemented in \textsc{JAX} and fully end-to-end compiled. The environment simulation and optimization processes are fused into a single computation graph, executed entirely on the target accelerator. To isolate execution performance, all reported times are measured after warming up the JIT compilation cache. All implementations and experiment configurations are contained within the supplementary experiments code.

\subsubsection*{Nonlinear Model Predictive Control}
For trajectory following, a standard NMPC formulation is implemented utilizing acados~\cite{verschueren2022acados}. The dynamics model used by the controller employs \textsc{Crazyflow}'s symbolic CasADi descriptions of the abstracted model, including thrust dynamics and drag terms. The optimization problem minimizes a standard least-squares cost function with weights $Q$ on the position error and $R$ on the input deviation from hover, subject to initial-state and input constraints. The optimal control problem consisting of $N$ nodes over a time horizon of $T$ is solved in a receding horizon manner with a control frequency of $f_\mathrm{ctrl}$. A summary of the controller parameters is provided in \autoref{tab:sup:mpc_params}.

\begin{table}
    \centering
    \caption{\textbf{NMPC parameters used for trajectory tracking.}}
    \begin{tabular}{l|l}
        Parameters          & Values    \\
        \hline 
        $Q$    & $80.0$    \\
        $R$   & $1.0$     \\
        $N$  & $25$      \\
        $T$                 & 2.5~s    \\
        $f_\mathrm{ctrl}$   & 100~Hz
    \end{tabular}
    \label{tab:sup:mpc_params}
\end{table}

\subsubsection*{Backpropagation Through Time}
The BPTT implementation computes analytical gradients directly through the unrolled simulation dynamics, requiring no value network for bootstrapping. Because analytical gradients exhibit lower variance than sampling-based gradient estimators, convergence is achieved using a significantly reduced number of parallel environments. To optimize for maximum training speed under these conditions, execution is performed entirely on the CPU, which processes lower environment counts more efficiently than the GPU. The environment rollouts and the BPTT optimization steps are fully fused into a single computation graph.

For the trajectory tracking task, environment reset positions are distributed along the reference trajectory. The drone is initialized at these starting points with hover thrust and the exact desired velocity for that point on the trajectory. We do not use any domain randomization, and the initial states are not randomized beyond the designated start positions along the trajectory. With the reward function
\begin{equation}
    \begin{split}
        r = {} & \exp\left(-2 \lvert p - p_\text{goal} \rvert\right) - w_\theta \lvert\theta\rvert - \sum_{i=1}^4 w_{\text{act},i}\left(a_i - a_{\text{hover},i}\right)^2 \\
               & - \sum_{i=1}^4 w_{\Delta\text{act},i} \left(a_i - a_{\text{prev},i}\right)^2,
    \end{split}
    \label{eq:trajectory_tracking_reward}
\end{equation}
where $w_\theta$, $w_{\text{act},i}$ and $w_{\Delta\text{act},i}$ are the weights of the angle reward, the action reward and the action smoothness reward, respectively, $\theta$ is the angle from the upright orientation, and $a_\text{hover}$ are the roll, pitch, yaw, and thrust commands for hovering, we obtain an end-to-end training time of 1.56~s for the 10~s trajectory. The list of hyperparameters for the BPTT setup is provided in \autoref{tab:sup:bptt_params}.

\begin{table}
    \centering
    \caption{\textbf{BPTT parameters for tracking the 10 s Lissajous.}}
    \begin{tabular}{l|l}
        Parameter           & Value                \\
        \hline
        Total timesteps     & $100{,}000$           \\
        $N_\mathrm{envs}$   & $16$                  \\
        Steps per rollout   & $40$                  \\
        Actor learning rate & $4.6 \times 10^{-2}$  \\
        Discount $\gamma$   & $1.0$                 \\
        Init logstd         & $-1.0$                \\
        $w_\theta$          & $0.1$                 \\
        $w_{\text{act}}$    & $[0.12, 0.12, 0.0, 0.02]$                 \\
        $w_{\Delta\text{act}}$    & $[1.4, 1.4, 0.0, 0.8]$           \\
    \end{tabular}
    \label{tab:sup:bptt_params}
\end{table}

For the parameters of the 5.5~s trajectory, the throwing application reward function, and its hyperparameters, we refer the reader to our supplementary experiment code.

\subsubsection*{Proximal Policy Optimization}
The PPO implementation utilizes Generalized Advantage Estimation (GAE) alongside entropy regularization. We use symmetric actor-critic observations, i.e., both networks receive identical observation vectors. The entire training pipeline is end-to-end compiled into a single \textsc{JAX} computation graph and executed exclusively on the GPU.

The continuous action space is normalized to the interval $[-1, 1]$, which maps linearly to the minimum and maximum allowable attitude angles and thrust commands. Following \cite{schuck2026primer}, we mitigate excessive initial training variance commonly associated with 3D rotational dynamics by setting the initial log standard deviation of the action distribution to $-1$.

Environment reset positions are handled as in the BPTT training, with no domain randomization or stochastic spatial perturbations applied to the initial states. PPO uses the same trajectory-tracking reward formulation as in \autoref{eq:trajectory_tracking_reward}, with separately tuned coefficients. A complete summary of the PPO hyperparameters is detailed in \autoref{tab:sup:ppo_params}.

\begin{table}
    \centering
    \caption{\textbf{PPO parameters for tracking the 10 s Lissajous.}}
    \begin{tabular}{l|l}
        Parameter                   & Value                \\
        \hline
        Total timesteps             & $2{,}000{,}000$       \\
        $N_\mathrm{envs}$           & $2048$                \\
        $N_\mathrm{minibatches}$    & $16$                  \\
        $N_\mathrm{epochs}$         & $10$                  \\
        Steps per rollout           & $16$                  \\
        Actor learning rate                    & $8 \times 10^{-4}$  \\
        Critic learning rate                   & $5 \times 10^{-3}$  \\
        Discount $\gamma$           & $0.92$               \\
        GAE lambda $\lambda$        & $0.94$               \\
        Init logstd                 & $-1.0$                \\
        Entropy coef.               & $7\times 10^{-3}$                \\
        $w_\theta$          & $0.1$                 \\
        $w_{\text{act}}$    & $[0.1, 0.1, 0.0, 0.02]$                 \\
        $w_{\Delta\text{act}}$    & $[0.9, 0.9, 0.0, 0.6]$           \\
    \end{tabular}
    \label{tab:sup:ppo_params}
\end{table}

\subsubsection*{Twin-Delayed DDPG}
The Twin-Delayed DDPG (TD3)~\cite{fujimoto18td3} experiments use the Crazyflie 2.1 platform equipped with brushed motors, which aligns with the experimental setup established in \cite{eschmann2024learning_to_fly_in_seconds}. We restrict TD3 training to a position-reaching task using direct rotor velocity commands and subsequently evaluate the resulting policies on trajectory-following tasks.

The target goal position for the reaching task is fixed. At environment reset, we randomize the initial Cartesian position, linear velocity, and angular velocity. As in PPO, we do not apply domain randomization during training. The maximum thrust-to-weight ratio for this Crazyflie is significantly below 2.0. We thus map the agent's continuous action space $(-1,1)$ to rotor velocity commands via a quadratic mapping, with \(0\) corresponding to hover rotor velocity and \(1\) to maximum rotor velocity. We set the minimum permitted rotor velocity above the physical minimum to keep the mapping monotonic.

Our implementation uses an asymmetric actor-critic architecture, providing the critic network with privileged information about the current rotor velocities. We also include a history of previous actions in the observation vectors. A complete summary of the TD3 hyperparameters is provided in \autoref{tab:sup:td3_params}.

\begin{table}
    \centering
    \caption{\textbf{TD3 parameters for hovering.}}
    \begin{tabular}{l|l}
        Parameter           & Value                \\
        \hline
        Total timesteps     & $20{,}000{,}000$       \\
        $N_\mathrm{envs}$   & $4096$                \\
        $N_\mathrm{epochs}$ & $8$                  \\
        $N_\mathrm{batch}$ & $2048$                  \\
        Memory size         & $4194304$                  \\
        Actor learning rate            & $1.34 \times 10^{-4}$  \\
        Critic learning rate           & $1.8 \times 10^{-3}$  \\
        Discount $\gamma$   & $0.984$               \\
        Polyak factor $\tau$ & $0.005$              \\
        Exploration noise $\sigma$ & $0.3$              \\
    \end{tabular}
    \label{tab:sup:td3_params}
\end{table}

\subsubsection*{BPTT Throw Policy}
We train a BPTT policy for throw recovery on the CPU using 32 parallel environments, and 180,000 environment steps. The policy outputs normalized thrust commands at 500 Hz. Episodes are initialized near the predicted takeover state with sampled perturbations to position, velocity, orientation, and angular velocity. The reward promotes recovery toward the desired height by penalizing height error, proximity to the floor and ceiling, orientation error, deviation from hover thrust, and action variation. In addition, it encourages low horizontal velocity and penalizes vertical motion directed away from the target height. A separate downward-velocity penalty is applied both above and below the target height, with a stronger penalty below it.

The policy is trained on a workstation with an Intel Core Ultra 9 285K and subsequently transferred to the Crazyflie over radio using a Crazyradio PA. The standard Crazyflie parameter protocol achieves approximately 600 parameter updates per second. To reduce the transfer time, we modified the firmware to batch multiple half-precision parameters into each packet and to omit the response packet for batched writes, increasing the effective transmission rate to approximately 8400 parameters per second. The policy is implemented as a neural network with 20 inputs, two hidden layers of 8 neurons each, and 4 outputs. All weights and biases are represented in half precision, giving a theoretical transfer time of approximately 33 ms. This compact architecture allows both policy training and transmission of the complete network to be completed within the 500 ms time budget of the experiment. The modified firmware with support for onboard neural-network inference is available at \url{https://github.com/learnsyslab/crazyflie-firmware}.

\subsubsection*{Model Predictive Path Integral}
The MPPI used in the obstacle avoidance showcase is implemented fully in \textsc{JAX}. Unlike in standard model predictive control, this formulation utilizes the full \textsc{Crazyflow} simulation as its internal forward-prediction model. The controller evaluates 500k parallel environments at a rate of 50~Hz. While the optimization can converge with fewer parallel rollouts, this configuration explicitly demonstrates the simulator's throughput and suitability for direct integration into real-time control loops.

Specifically, we use a hybrid implementation of iCEM~\cite{pinneri21icem} and MPPI~\cite{williams2017mppi} with explicit modes, running five independent controllers in parallel and selecting the best control sequence at each step. Following iCEM, each controller truncates the top 1\% of trajectories, uses temporally correlated noise, and warm-starts its sampling from the previous mean solution. To obtain the mean across the best samples, we use exponential weighting over all elites, as in MPPI. The full set of parameters can be found in our experiments code.

\subsubsection*{Vision-based Flight} Our vision-based experiment uses \textsc{Crazyflow}'s Gaussian splatting capabilities to train an end-to-end, zero-shot deployable policies that navigate the drone to desired goal positions. In our experiments, we record 10 minutes of standard phone video footage, and reconstruct the flight space from it. Each reconstruction requires less than four minutes of training time, and training vision-based policies on the reconstructions of our flight space takes approximately 14 minutes. The full scan-to-fly pipeline including manual work (e.g., filming) requires less than 1.5 hours.

The policy is trained with a DAgger-like setup~\cite{ross2011dagger}. We sample random start- and goal positions, compute an expert signal using the state-based Mellinger controller, extract the desired attitude command, and distill that to the vision-based student policy using an L1 loss. The student action is then executed, except for 10\% of the environments, which execute the expert commands for additional state space coverage.

The policy consists of a four-layer convolutional encoder for the image input, a trunk for joint processing of image embeddings and onboard information, and two prediction heads. The input stacks four 64x64 grayscale images to give the policy an additional sense of velocities, and fuses the embedding with sensor readings from the onboard gyro and accelerometer, the previous command, and the goal in the world frame. The policy outputs normalized attitude and collective thrust commands, and a prediction of its current position, orientation, velocity, and angular velocity at 50 Hz. We use the state predictions in an auxiliary state estimation loss that is critical for stable training.

To make the policy robust against visual differences and sensor noise, we use flight disturbances, sensor bias, drift and noise, randomize the camera position in each world, and vary brightness, contrast etc. We achieve further robustness against smaller structural differences by training on an ensemble of splats that reconstruct the same space. A full list of training parameters can be found in \autoref{tab:sup:vision_params}.

\begin{table}
    \centering
    \caption{\textbf{DAgger parameters for vision-based flight.}}
    \begin{tabular}{l|l}
        Parameter           & Value                \\
        \hline
        Total timesteps     & $51{,}200{,}000$       \\
        $N_\mathrm{envs}$   & $1024$                \\
        $N_\mathrm{epochs}$ & $500$                  \\
        $N_\mathrm{batch}$ & $256$                  \\
        Memory size         & $1{,}048{,}576$                  \\
        Learning rate            & $3\times 10^{-4}$, cosine decay  \\
        Frame drops & $5\%$                  \\
        Camera latency & $20-60$ ms                  \\
        Action latency & $0-20$ ms                  \\
        Camera mount & $\pm5$ mm, $\pm0.04$ rad, $\pm0.04^{\circ}$ FOV                 \\
        
    \end{tabular}
    \label{tab:sup:vision_params}
\end{table}

\end{document}